\RequirePackage[svgnames]{xcolor}

\documentclass[11pt,a4paper]{mystyle}

\usepackage[all]{hypcap}
\usepackage[svgnames]{xcolor}
\usepackage[comma,authoryear,compress]{natbib}
\usepackage{algorithm}
\usepackage{algorithmicx}
\usepackage{algpseudocode}
\usepackage{microtype}
\usepackage{booktabs}
\usepackage{float}
\usepackage{bigstrut}
\usepackage{multirow}
\usepackage{bm}
\usepackage{dsfont}
\usepackage{enumitem}
\usepackage{graphicx}
\usepackage{cleveref}
\usepackage{bxcoloremoji}

\usepackage{float}

\usepackage[utf8]{inputenc} %
\usepackage[T1]{fontenc}    %
\usepackage{hyperref}       %
\usepackage{url}            %
\usepackage{booktabs}       %
\usepackage{amsfonts}       %
\usepackage{nicefrac}       %
\usepackage{microtype}      %
\usepackage{amssymb}
\usepackage{fdsymbol}
\usepackage{wrapfig}
\usepackage{lipsum}
\usepackage{enumitem}
\usepackage{stackengine}
\usepackage[font=small,labelfont=bf]{caption}
\usepackage{color}
\usepackage{xcolor}
\usepackage{adjustbox}

\usepackage{rotating}
\usepackage{makecell}
\usepackage{colortbl}
\usepackage{tcolorbox}
\usepackage{subfigure}
\usepackage{multirow}
\usepackage{pifont}

\definecolor{cell_high}{rgb}{0.878, 0.949, 0.835}

\title{Enhancing the Comprehensibility of Text Explanations via Unsupervised Concept Discovery}

\runningtitle{Enhancing the Comprehensibility of Text Explanations via Unsupervised Concept Discovery}

\author[1,2]{\href{https://scholar.google.com/citations?user=sv7uxi4AAAAJ}{\textcolor{black}{Yifan Sun}}}
\author[1]{\href{https://scholar.google.com/citations?user=hGZwK0cAAAAJ}{\textcolor{black}{Danding Wang}}}
\author[1]{\href{https://sheng-qiang.github.io/}{\textcolor{black}{Qiang Sheng}}}
\author[1,2]{\href{https://scholar.google.com/citations?user=fSBdNg0AAAAJ}{\textcolor{black}{Juan Cao}}}
\author[1,2]
{\href{mailto:jtli@ict.ac.cn}
{\textcolor{black}{Jintao Li}}}

\affil[1]{Media Synthesis and Forensics Lab, Institute of Computing Technology, Chinese Academy of Sciences}
\affil[2]{University of Chinese Academy of Sciences}

\correspondingauthor{Danding Wang}
\emails{{\{\href{mailto:sunyifan23z@ict.ac.cn}{sunyifan23z},\href{mailto:wangdanding@ict.ac.cn}{wangdanding},\href{mailto:shengqiang18z@ict.ac.cn}{shengqiang18z},\href{mailto:caojuan@ict.ac.cn}{caojuan},\href{mailto:jtli@ict.ac.cn}{jtli}\}@ict.ac.cn}}

\begin{document}

\begin{abstract}
Concept-based explainable approaches have emerged as a promising method in explainable AI because they can interpret models in a way that aligns with human reasoning. However, their adaption in the text domain remains limited. Most existing methods rely on predefined concept annotations and cannot discover unseen concepts, while other methods that extract concepts without supervision often produce explanations that are not intuitively comprehensible to humans, potentially diminishing user trust. These methods fall short of discovering comprehensible concepts automatically. 
To address this issue, we propose \textbf{ECO-Concept}, an intrinsically interpretable framework to discover comprehensible concepts with no concept annotations. 
ECO-Concept first utilizes an object-centric architecture to extract semantic concepts automatically. Then the comprehensibility of the extracted concepts is evaluated by large language models. Finally, the evaluation result guides the subsequent model fine-tuning to obtain more understandable explanations.
Experiments show that our method achieves superior performance across diverse tasks. 
Further concept evaluations validate that the concepts learned by ECO-Concept surpassed current counterparts in comprehensibility.
\vspace{5mm}

\coloremojicode{1F4C5} \textbf{Date}: May 26, 2025

\coloremojicode{1F3E0} \textbf{Project}: \href{https://vicki-sun.github.io/projects/ECO-Concept}{https://vicki-sun.github.io/projects/ECO-Concept}

\coloremojicode{1F4AC} \textbf{Venue}: ACL 2025 Findings 

\end{abstract}

\maketitle
\vspace{3mm}

\begin{figure}[ht]
\centering
\includegraphics[width=0.6\textwidth]{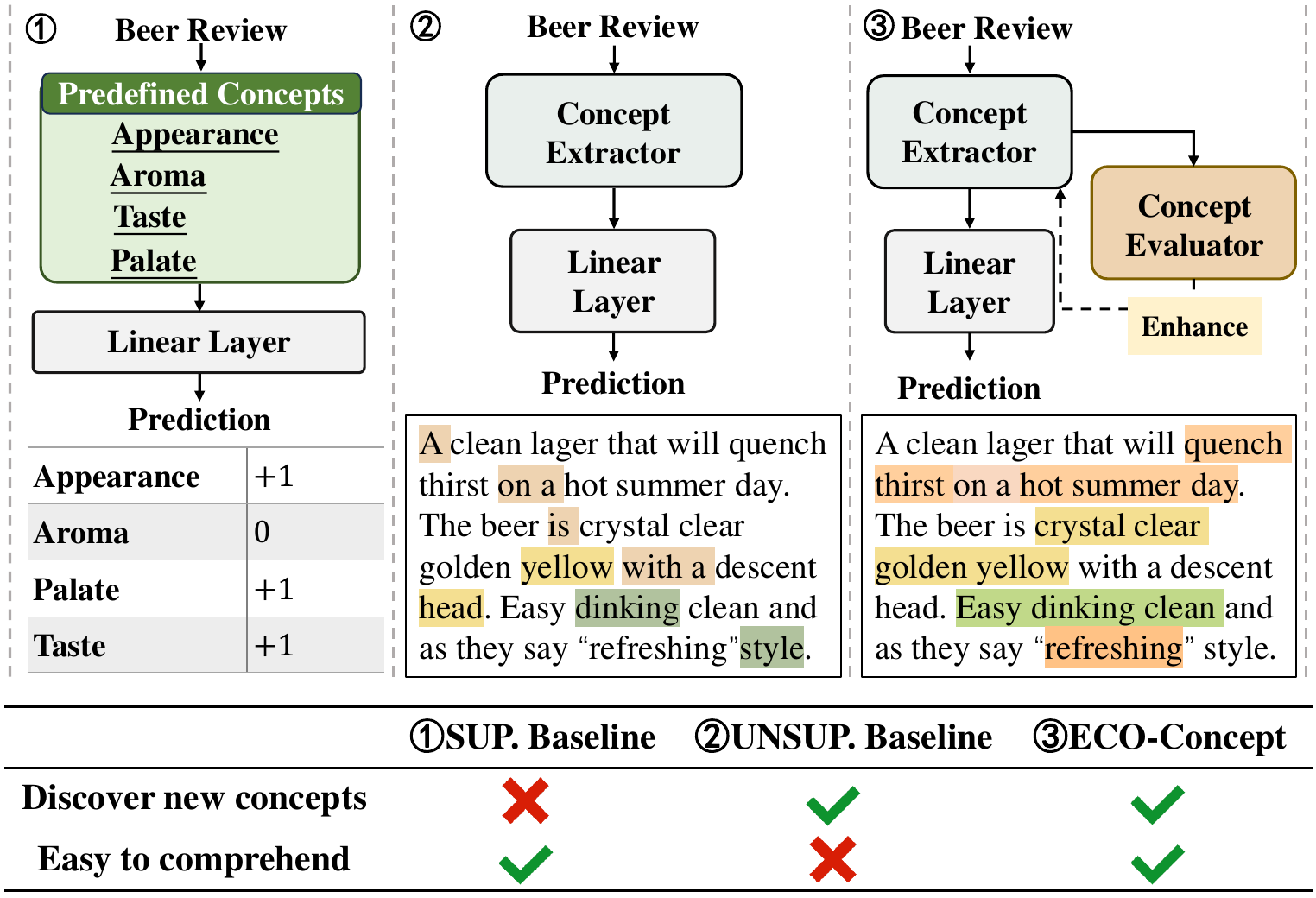}
\caption{Comparison of explanations between our proposed \scalebox{1.2}{\ding{194}}~ECO-Concept and existing typical \scalebox{1.2}{\ding{192}}~supervised and \scalebox{1.2}{\ding{193}}~unsupervised concept-based methods. Supervised methods explain based on predefined concepts, while ECO-Concept and unsupervised methods explain via concept-related highlighted text. ECO-Concept eliminates the need for concept annotations and can discover unseen concepts with improved comprehensibility.}
\label{fig:intro}
\end{figure}

\section{Introduction}
\label{sec:introduction}
Deep neural language models lack explainability. 
A recent way to tackle this issue is concept-based explanations~\citep{achtibat2023attribution,Poeta2023ConceptbasedEA}, which map the inputs to a set of concepts and measure the importance of each concept to model predictions. By offering human-understandable attributes, concept-based explanations better resemble the way humans reason and explain. 

Some existing concept-based methods apply post-hoc analysis to a trained model, providing concepts that either explain a model's prediction~\citep{kim2018interpretability,goyal2019explaining,ghorbani2019towards,jourdan2023cockatiel} or elucidate its internal network structure~\citep{dalvi2022discovering,sajjad2022analyzing,bills2023language,bricken2023towards}. Since the model was not exposed to these concepts during training, it may lead to unfaithful interpretations or meaningless concept activations~\citep{rudin2019stop}.
Self-explaining concept-based methods offer a more reliable approach by providing built-in explanations along with their predictions. However, most of these methods require abundant annotations~\cite{tan2024sparsity,tan2024interpreting,de2024self} for predefined concepts, which shows their limitations in automatically discovering new concepts during training. 
Though there exist self-explaining methods that extract concepts without concept annotations~\citep{rajagopal2021selfexplain,antognini2021rationalization,das2022prototex}, the extracted concepts often lack human comprehensibility due to the ignorance of such constraints to guide concept extraction. 
Consequently, the resulting concept explanations may highlight irrelevant or misleading information, bringing more confusion (demonstrated by previous research~\citep{chaleshtori2024evaluating}) and less trust in humans (revealed by our human forward simulatability experiments).
The comparison is illustrated in Fig.~\ref{fig:intro}.

In this paper, we aim to build a model that can automatically discover comprehensible concepts with no need for concept supervision. Inspired by the recent application of the slot attention mechanism~\citep{locatello2020object} in concept-based interpretability for vision tasks~\citep{wang2023learning,hong2024concept}, we develop a concept extractor that leverages slot attention, enabling each slot to learn a distinct task-specific concept. 
To ensure that the learned concepts are human-understandable, we leverage LLMs as human proxies to evaluate concept comprehensibility during model training and use the results to refine the concept extractor. 
Specifically, we propose a metric for concept comprehensibility that measures the capability of a concept's activation map to be summarized and reconstructed using natural language.
Building on prior work that employs large language models to generate neuron explanations and simulate activations~\citep{bills2023language,templeton2024scaling}, our approach introduces a novel feedback loop. This loop utilizes comprehensibility scores to refine the concept extractor, making the extracted concepts more intuitive and human-understandable.
Experiments on seven tasks and human studies both show the superiority of our method compared with existing concept-based explanation methods. Our contributions are summarized: 

\begin{itemize}
[nosep,leftmargin=1em,labelwidth=*,align=left]
    \item We propose a method to evaluate the human interpretability of concepts by employing LLMs as human proxies, enabling real-time assessment of concept comprehensibility during training.  
    \item We design a self-explaining model training mechanism with no concept annotations, where LLMs evaluate the discovered concepts and guide the subsequent model to learn concepts with greater interpretability.
    \item Our method demonstrates performance comparable to black-box models across various tasks, while the learned concepts are interpretable and comprehensible to humans.
\end{itemize}

\section{Related Work}
\label{sec:related_work}
\textbf{Concept-based Post-hoc Methods} explain an existing model without modifying its internal architecture. Some studies predefine task-related concepts and assess them by quantifying contributions with linear probes~\citep{alvarez2018towards,crabbe2022concept} or analyzing causal effects through proxy methods~\citep{goyal2019explaining,wu2023causal}. Methods without concept supervision extract concepts from intermediate layer representations using K-Means~\citep{ghorbani2019towards}, Non-Negative Matrix Factorization~\citep{zhang2021invertible, fel2023craft, jourdan2023cockatiel} or concept completeness maximization~\citep{yeh2020completeness}.

However, these post-hoc methods cannot guarantee that models truly comprehend or employ the adopted concepts, as the models were not exposed to these concepts during training~\citep{rudin2019stop,Poeta2023ConceptbasedEA}. Additionally, some unsupervised post-hoc methods may extract concepts lacking semantic meaning.

\noindent
\textbf{Concept-based Self-explaining Methods}
aim to provide a built-in human-interpretable explanation along with the prediction. Some methods adopt a supervised paradigm with experts manually crafting a set of concepts. A representative framework is Concept Bottleneck Model~(CBM)~\citep{koh2020concept}, where an intermediate concept bottleneck layer is introduced to break the standard end-to-end training paradigm. 
Recently, some studies have taken advantage of the capabilities of LLMs to generate concepts for each class as a replacement for manual annotation~\citep{yang2023language, oikarinen2023label}. This insight has also been applied in the text domain~\citep{tan2024interpreting, de2024self}. 
These methods rely on concepts that are predefined based on priors and cannot discover or adapt to new concepts during training. An incomplete or biased predefined concept set could severely compromise both the model's interpretability and performance.
Text Bottleneck Model~(TBM)~\citep{ludan2023interpretable} utilized LLMs to automatically discover and measure concepts. However, in TBM, the mapping from samples to concepts is directly determined by LLMs,  which remain largely opaque and function as a black box.

Unsupervised self-explaining concept models autonomously extract concepts during model training without the need for concept annotations. SENN~\citep{alvarez2018towards} uses self-supervision by reconstruction loss for concept discovery. BotCL~\citep{wang2023learning} and CCTs~\citep{hong2024concept} utilize a slot attention-based mechanism to extract task-dependent concept slots. While in the text domain, SelfExplain~\citep{rajagopal2021selfexplain} identifies concepts as the non-terminal leaves of the semantic tree parsing the text. Some methods classify samples based on their distance to learnable prototypes~\citep{li2018deep, das2022prototex, xie2023proto} and introduce distance-based losses to guide prototype learning. 
Without concept annotations, the concepts extracted by these methods pose challenges to human understanding. Although existing methods evaluate human interpretability through manual assessments during the experimental phase, no approach has yet incorporated human feedback into the training to optimize concept comprehensibility.

\section{Method}
\label{sec:method}

\begin{figure*}[!t]
\centering
\includegraphics[width=\textwidth]{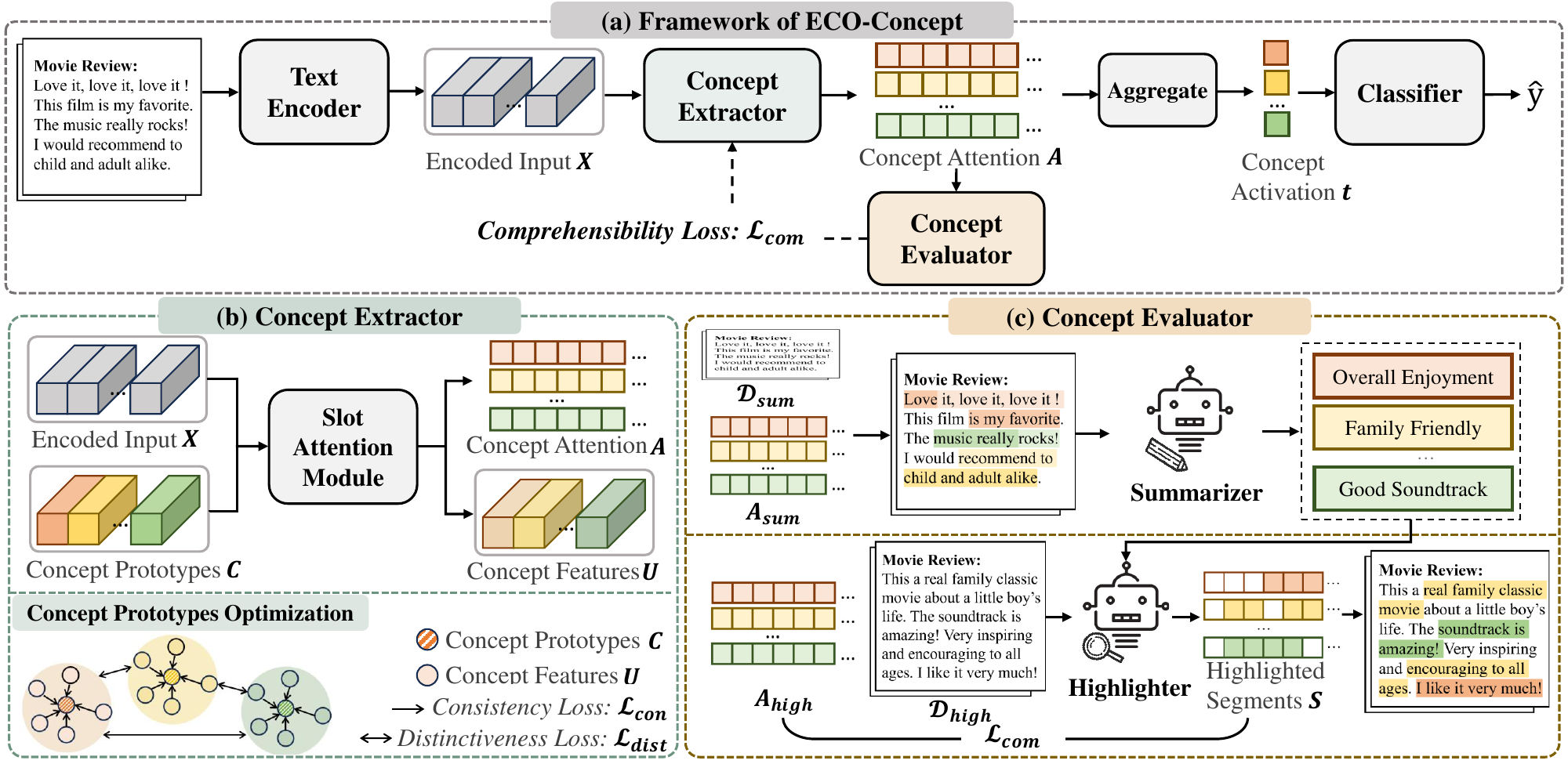}
\caption{(a)~Illustration of the proposed framework ECO-Concept. ECO-Concept consists of a concept extractor, a classifier, and a concept evaluator. (b)~The concept extractor takes the encoded text $\bm{X}$ as input and interacts with the concept prototypes $\bm{C}$ to obtain a slot attention matrix $\bm{A}$ and concept features $\bm{U}$. The concept prototypes are optimized using consistency and distinctiveness loss. (c)~The concept evaluator utilizes exemplars with the highest concept attention values to construct two sets, $\mathcal{D}_{sum}$ and $\mathcal{D}_{high}$. Using the corresponding slot attention matrices $\bm{A}_{sum}$ and $\bm{A}_{high}$, the evaluator highlights these exemplars to perform concept summarization and highlighting, thus getting the comprehensibility loss to guide model fine-tuning.}
\label{fig:framework}
\vspace{-0.2cm}
\end{figure*}

\subsection{Framework}
Given a document consists of $L$ tokens~(words), we adopt a text encoder to encode tokens of the document as $\bm{X}=\{\bm{x}_1, \bm{x}_2, ..., \bm{x}_L\} \in \mathbb{R}^{L \times D}$, where $D$ is the dimension of the encoded space. ECO-Concept learns a set of $M$ concept prototypes $\bm{C}=\{\bm{c}_1, \bm{c}_2, ..., \bm{c}_M\} \in \mathbb{R}^{M \times D}$while learning the original classification task.
Each concept prototype $\bm{c}_m$ is a trainable $D$-dimensional variable and is initialized randomly. These prototypes are continuously optimized during the training process to represent task-relevant concepts.
Fig.~\ref{fig:framework} shows an overview of our framework. ECO-Concept consists of three modules: a concept extractor, a classifier, and a concept evaluator. The concept extractor takes the encoded text as input and generates a concept slot attention matrix by interacting with the concept prototypes. The slot attention scores across tokens are then summed and fed into the classifier for prediction. To enhance the comprehensibility of the concepts, we designed a concept evaluator that receives slot attention from the concept extractor, maps them back to the original text, and uses human proxies~(here, LLMs)\footnote{We conducted a human assessment of using LLMs to evaluate concept comprehensibility. The details are in Appendix~\ref{app:human_verification}.} to summarize the concepts, and highlights concept-related segments with additional exemplars. If a concept is easy to understand, the highlighted segments should be similar to the model’s slot attention. We use the difference between the highlighted segments and the model’s slot attention scores, considering the importance of each concept, as a comprehensibility loss. This feedback is used to make the learned concepts easier for human understanding.

\subsection{Concept Extractor}
Concept Extractor uses slot attention to discover concepts automatically. Slot Attention~\citep{locatello2020object} was initially applied in object recognition, with its core mechanism focusing on utilizing slots to compete for explaining parts of the input features. Through multiple rounds of attention competition, these slots are gradually adjusted to represent distinct object features, i.e., the underlying concepts. In computer vision, several methods already utilize the slot attention mechanism to achieve concept-based interpretability~\citep{wang2023learning, hong2024concept}. Inspired by these approaches, we implemented a concept extractor based on slot attention, where each slot learns a distinct task-dependent concept. 

Our concept extractor takes the encoded features $\bm{X}$ as input and interacts with the concept prototypes $\bm{C}$ via the slot attention module, obtaining the concept features $\bm{U} \in \mathbb{R}^{M \times D}$ and the slot attention matrix $\bm{A} \in [0, 1)^{M \times L}$. 
For this, we apply linear projection $\bm{W}_q$ on the concept slots to obtain the queries and projections $\bm{W}_k$ and $\bm{W}_v$ on the inputs to obtain the keys and the values, all having the same dimension $D$. Then, we perform a dot product between the queries and keys to get the attention matrix $\bm{A}$.  
{\setlength{\abovedisplayskip}{5pt}   
\setlength{\belowdisplayskip}{5pt}
\begin{equation}
\label{eq:slot_attention}
\bm{A}=\phi(\frac{(\bm{W}_q\bm{C})(\bm{W}_k\bm{X})^T}{\sqrt{D}}). 
\end{equation}}

In the slot attention matrix $\bm{A}$, each element $\bm{A}_{m,l}$ represents the attention weight of the concept slot $m$ when attending to the input vector $l$. Unlike traditional attention mechanisms, we normalize $\bm{A}$ by applying a sparse softmax function $\phi$ across the concept slots~(along the $M$ axis). This normalization introduces competition among slots to attend to each input token. The sparsity normalization ensures that each input token is primarily associated with a limited number of concepts, facilitating a more focused and interpretable learning of conceptual representations.
We then aggregate features in $\bm{X}$ corresponding to each concept into concept features $\bm{U} \in \mathbb{R}^{M \times D}$.
{\setlength{\abovedisplayskip}{5pt}   
\setlength{\belowdisplayskip}{5pt}
\begin{equation}
\label{eq:slot_feature}
\bm{U}=\frac{\bm{A}}{\sum_l^L \bm{A}_{:, l}} (\bm{W}_v\bm{X}).
\end{equation}}

For better interpretability, we employ concept regularizers to constrain the training of concept prototypes $\bm{C}$. 

\noindent
\textbf{Consistency} The extracted concepts should represent consistent semantics. That is, the concept features activated by each concept prototype across the samples should not have large variations. The concept features $\bm{u}_m$ and $\bm{u}'_m$ under the same concept $m$ of different documents should be similar to each other. Under each concept, we select the top $k$ samples with the highest activation values in a mini-batch. We define the consistency loss as:
{\setlength{\abovedisplayskip}{5pt}   
\setlength{\belowdisplayskip}{5pt}
\begin{equation}
\label{eq:consistency_loss}
\mathcal{L}_{con} = \frac{1}{M}\sum_m \sum_{\bm{u}_m,\bm{u}'_m}\frac{\|\bm{u}_m - \bm{u}'_m\|_2^2}{k(k-1)}.
\end{equation}}

\noindent
\textbf{Distinctiveness} To capture different aspects of documents, different concepts should cover different elements. This means that the average features of concept $m$ calculated based on the top $k$ samples with the highest activation values in a mini-batch, given by $\displaystyle \bar{\bm{u}}_m = (1/k)\sum \bm{u}_m$, should be different from any other. We encode this into a loss as:
{\setlength{\abovedisplayskip}{5pt}   
\setlength{\belowdisplayskip}{5pt}
\begin{equation}
\label{eq:distinctiveness_loss}
\mathcal{L}_{dist} = -\sum_{\bar{\bm{u}}_m,\bar{\bm{u}}_{m'}}\frac{\|\bar{\bm{u}}_m - \bar{\bm{u}}_{m'}\|_2^2}{k(k-1)}.
\end{equation}}

\subsection{Classifier}
We use a fully connected layer for classification, and total concept activation $\bm{t} \in \mathbb{R}^M$ is the only input as the concept bottleneck~\citep{koh2020concept}:
{\setlength{\abovedisplayskip}{5pt}   
\setlength{\belowdisplayskip}{5pt}
\begin{equation}
\label{eq:concept_activation}
\bm{t} = \sum_{l=1}^{L}\bm{A}_{:,l}.
\end{equation}}
Formally, letting $\bm{W} \in \mathbb{R}^{M \times \Omega}$ be a learnable matrix, prediction $\hat{y}$ is given by:
{\setlength{\abovedisplayskip}{5pt}   
\setlength{\belowdisplayskip}{5pt}
\begin{equation}
\label{eq:prediction}
\hat{y} = \bm{W}\bm{t}.
\end{equation}}

We use softmax cross-entropy for the classification task, donated by $\mathcal{L}_{ce}$. The overall loss is
{\setlength{\abovedisplayskip}{5pt}   
\setlength{\belowdisplayskip}{5pt}
\begin{equation}
\label{eq:stage1_loss}
\mathcal{L} = \mathcal{L}_{ce} + \lambda_{con}\mathcal{L}_{con}+ \lambda_{dist}\mathcal{L}_{dist},
\end{equation}}
where $\lambda_{con}$ and $\lambda_{dist}$ are weight terms.

\subsection{Concept Comprehensibility Enhancement}
By optimizing the above loss function, our model achieved effective classification performance while extracting task-relevant concepts with semantic meanings. However, due to the lack of supervision from concept annotations, the identified concepts may still exhibit some degree of semantic ambiguity. To address this, we aim to further enhance the comprehensibility of the extracted concepts, prioritizing those of higher importance for improved human interpretability. Specifically, we evaluate each concept's importance and comprehensibility, combining these metrics accordingly to define a comprehensibility loss. This loss is integrated into the original loss function, and the model is further fine-tuned to improve concept comprehensibility.

\noindent\textbf{Concept Importance.}
The weight matrix of the classifier $\bm{W}$ learns the correlation between the class and concepts. For each concept $m$, $\bm{W}_{m,\omega}$ is the correlation from concept $m$ to class $\omega$. A positive value of $\bm{W}_{m,\omega}$ means that concept $m$ co-occurs with class $\omega$ in the dataset, so its presence positively supports class $\omega$. Meanwhile, a negative value means the concept $m$ and the class $\omega$ rarely co-occurs. When concept $m$ is present, the class is unlikely to be $\omega$. Regardless of whether the value is positive or negative, $\bm{W}_{m,\omega}$ represents the relationship between concept $m$ and class $\omega$. Therefore, we take the absolute value of $\bm{W}_{m,\omega}$ here and compute the sum of the absolute values of all the weights connecting concept $m$ as $\sum_{\omega = 1}^{\Omega} | \bm{W}_{m,\omega} |$.

The importance of concept $m$ should also take into account the concept activation. Given the average concept activation $\bm{t}_m$ in a mini-batch $\mathcal{B}$, the concept importance score $\beta_m$ is:
{\setlength{\abovedisplayskip}{5pt}   
\setlength{\belowdisplayskip}{5pt}
\begin{equation}
\label{eq:concept_importance}
\beta_m = \frac{1}{|\mathcal{B}|} \sum_{\bm{t} \in \mathcal{B}} \bm{t}_m
\sum_{\omega = 1}^{\Omega} | \bm{W}_{m,\omega} |.
\end{equation}}

\noindent\textbf{Concept Comprehensibility.} To measure the comprehensibility of concepts, we refer to the automated interpretability evaluation methods used for explaining neurons in LLMs~\citep{bills2023language, templeton2024scaling}. In summary, the auto-interpretability procedure takes samples of text where the neurons activate, asks a language model to write a human-readable interpretation of the neuron features, and then prompts the language model to use
this description to predict the neurons' activation on other samples of text. The correlation between the model’s predicted and the actual activations is the feature’s interpretability score. 

We adopt a similar idea, measuring the comprehensibility of a concept by evaluating its ability to be summarized and reconstructed using natural language. 
For each concept, we select the exemplars with the highest attention values to form two sets, $\mathcal{D}_{sum}$ and $\mathcal{D}_{high}$, which are used for concept summarization and concept-related segment highlightings, respectively. We first present the exemplars $\mathcal{D}_{sum}$ with their corresponding slot attention values and ask the LLM to generate the interpretation. If it assesses that this concept has semantic meanings, we then prompt another LLM to highlight concept-related tokens on new exemplars from $\mathcal{D}_{high}$ in a 0-1 scale. For each highlighted exemplar, its slot attention matrix is $\bm{A} \in [0, 1)^{M \times L}$ and the highlighted matrix obtained by LLM is $\bm{S} \in [0, 1]^{M \times L}$. For concepts that are considered semantically meaningless by LLM, our goal is to minimize their corresponding activations. Specifically, we achieve this by setting the elements of $\bm{S}$ to zero for highlighted exemplars corresponding to these concepts.

The comprehensibility score of each concept is obtained by averaging the MSE loss between the slot attention matrix $\bm{A}$ and the highlighted matrix $\bm{S}$ for the highlighted samples within a mini-batch $\mathcal{B}$. To ensure that more important concepts are easier to understand, the comprehensibility loss is computed as a weighted combination of concept importance and comprehensibility:
{\setlength{\abovedisplayskip}{5pt}   
\setlength{\belowdisplayskip}{5pt}
\begin{equation}
\label{eq:simulation_loss}
\mathcal{L}_{com}\!=\!\frac{1}{M}\! \sum_{m=1}^M\! \frac{\beta_m \sum_{\bm{A} \in \mathcal{B}_m \cap \mathcal{D}_{high}}\|\bm{A}_{m, :}\!-\!\bm{S}_{m, :}\|_2^2}{|\mathcal{B}_m \cap \mathcal{D}_{high}|}\!.
\end{equation}}

\noindent\textbf{Training Strategies.}
Based on the model trained in the first phase, we continue to train the concept prototypes $\bm{C}$ and the classifier using the following loss function:
{\setlength{\abovedisplayskip}{5pt}   
\setlength{\belowdisplayskip}{5pt}
\begin{equation}
\label{eq:stage2_loss}
\mathcal{L}\!=\!\mathcal{L}_{ce}\!+\! \lambda_{con}\mathcal{L}_{con}\!+ \!\lambda_{dist}\mathcal{L}_{dist}\!+ \!\lambda_{com}\mathcal{L}_{com}.
\end{equation}}

At the end of each iteration, we conduct a concept re-summarization. If the meaning of a concept remains unchanged, the corresponding concept prototype parameters are frozen in the following iterations. Otherwise, we highlight concept-related segments based on the updated summarization. This process is repeated until the meanings of all concepts stabilize.

\section{Experiments}
\label{sec:experiments}
In this section, we conduct extensive experiments to verify the task performance and interpretability of ECO-Concept. Due to space limitations, parameter sensitivity analysis and other experiments are included in Appendices~\ref{app:sensitivity} and \ref{app:robustness}.

\subsection{Experimental Settings}
\noindent\textbf{Datasets}
We conduct experiments on seven public datasets. To compare our results with supervised methods, we utilize three datasets with concept annotations: CEBaB~\citep{abraham2022cebab}, Hotel~\citep{wang2010latent}, and Beer~\citep{mcauley2012learning}. The rest four datasets, IMDB~\citep{maas2011learning}, AGnews~\citep{gulli2004ag}, Twitter~\citep{sheng2021integrating}, and SciCite~\citep{cohan2019structural}, do not include concept annotations. More details about these datasets are in Appendix~\ref{app:datasets}.

\noindent\textbf{Baselines}
The selected baselines include black-box, supervised concept-based, and unsupervised concept-based methods. The first group, black-box methods, directly tackles text classification tasks without interpretatability, including a \textbf{BERT-based classifier}~\citep{devlin2018bert} and a \textbf{RoBERTa-based classifier}~\citep{liu2019roberta}. The second group, supervised concept-based methods, leverages concept annotations to predict both the presence of concepts and the target class, including \textbf{CBM}~\citep{kim2018interpretability} and an evolved CBM variant \textbf{SparseCBM}~\citep{tan2024sparsity}. The third group, unsupervised concept-based methods, comprise two self-explaining methods \textbf{SelfExplain}~\citep{rajagopal2021selfexplain} and \textbf{PROTOTEX}~\citep{das2022prototex}, and two post-hoc methods \textbf{COCKATIEL}~\citep{jourdan2023cockatiel} and \textbf{Concept-Shap}~\citep{yeh2020completeness}. The details of these methods are included in Appendix~\ref{app:baselines}.

\noindent\textbf{Implementation Details}
For all concept-based methods, we use RoBERTa as the text encoder and set the number of concepts to 20 across all tasks. For ECO-Concept, we provide the top 10 exemplars for each concept and utilize GPT-4o to summarize the concept interpretation. Additionally, we prompt GPT-4o-mini to highlight text segments on 100 exemplars per concept. 
The trade-off parameters $\lambda_{con}$, $\lambda_{dist}$, and $\lambda_{com}$ are 0.1, -0.01, and 1, respectively. In our experiments, the concept comprehensibility enhancement stage ends within three iterations.
We run all the experiments on an NVIDIA A800 GPU with 80GB RAM.

\subsection{Task Performance}
Table~\ref{tab:task_performance} shows the performance comparison of ECO-Concept and baselines. In general, our method achieves superior classification performance across various datasets.
Compared to supervised methods, ECO-Concept achieves competitive results with no concept supervision, indicating its ability to automatically discover new concepts without compromising performance. Compared to unsupervised baselines, ECO-Concept shows significant improvements in both accuracy and F1 with pairwise t-tests at a 95$\%$ confidence level, validating the effectiveness of its conceptual representations. Moreover, ECO-Concept also has comparable or better performance compared with black-box models. This indicates that it effectively balances both task-discriminativity and concept comprehensibility, showing the potential to build interpretable models without performance trade-offs.
\begin{table*}[ht]
\centering
\caption{Classification performance of ECO-Concept and other baselines. The best result is \colorbox{cell_high}{highlighted}.}
\resizebox{\textwidth}{!}{
\begin{tabular}{@{}c|c| cc cc cc cc cc cc cc@{}}
\toprule
\multirow{2}{*}{\textbf{Category}} & \multirow{2}{*}{\textbf{Method}} & \multicolumn{2}{c}{\textbf{CEBaB}} & \multicolumn{2}{c}{\textbf{Beer}} & \multicolumn{2}{c}{\textbf{Hotel}} & \multicolumn{2}{c}{\textbf{IMDB}} & \multicolumn{2}{c}{\textbf{AGnews}} & \multicolumn{2}{c}{\textbf{Twitter}} & \multicolumn{2}{c}{\textbf{SciCite}} \\
\cmidrule(lr){3-4} \cmidrule(lr){5-6} \cmidrule(lr){7-8} \cmidrule(lr){9-10} \cmidrule(lr){11-12} \cmidrule(lr){13-14} \cmidrule(lr){15-16}
 &  & Acc & F1 & Acc & F1 & Acc & F1 & Acc & F1 & Acc & F1 & Acc & F1 & Acc & F1 \\
\midrule
\multirow{2}{*}{Black-Box} & RoBERTa & .682 & .797 & .882 & .882 & \cellcolor{cell_high} \textbf{.981} & \cellcolor{cell_high} \textbf{.981} & \cellcolor{cell_high} \textbf{.937} & \cellcolor{cell_high} \textbf{.937} & \cellcolor{cell_high} \textbf{.941} & .960 & \cellcolor{cell_high} \textbf{.828} & .812 & .858 & .879 \\
 & BERT & .640 & .770 & .878 & .878 & .975 & .975 & .912 & .912 & .934 & .956 & .775 & .745 & .845 & .861\\
\midrule
\multirow{2}{*}{SUP.} & CBM & .669 & .802 & .883 & \cellcolor{cell_high} \textbf{.885} & .979 & .979 & - & - & - & - & - & - & - & - \\
 & SparseCBM & .644 & .767 & .883 & .883 & .981 & .979 & - & - & - & - & - & - & - & - \\
\midrule
\multirow{3}{*}{UNSUP.} & SelfExplain & .683 & .799 & .873 & .872 & .978 & .979 & .936 & .936 & .925 & .949 & .817 & .806 & .856 & .873 \\
 & PROTOTEX & .610 & .728 & .877 & .877 & .977 & .977 & .935 & .935 & .915 & .943 & .826 & .812 & .852 & .871 \\
 & ECO-Concept& \cellcolor{cell_high} \textbf{.697} & \cellcolor{cell_high} \textbf{.808} & \cellcolor{cell_high} \textbf{.885} & \cellcolor{cell_high} \textbf{.885} & \cellcolor{cell_high} \textbf{.981} & \cellcolor{cell_high} \textbf{.981} & \cellcolor{cell_high} \textbf{.937} & \cellcolor{cell_high} \textbf{.937} & \cellcolor{cell_high} \textbf{.941} & \cellcolor{cell_high} \textbf{.961} & \cellcolor{cell_high} \textbf{.828} & \cellcolor{cell_high} \textbf{.813} & \cellcolor{cell_high} \textbf{.860} & \cellcolor{cell_high} \textbf{.881}\\
\bottomrule
\end{tabular}}
\label{tab:task_performance}
\vspace{-0.2cm}
\end{table*}

\subsection{Concepts Evaluation}
To evaluate the comprehensibility of the concepts extracted by our method, we first define three quantitative metrics. Then we conduct several human evaluations to further assess how easily these concepts can be understood. 
We compare our method's extracted concepts with those from three other global unsupervised concept extraction methods. Among these methods, Cockatiel and Concept-Shap are post-hoc methods, and ProtoTEx is a self-explaining method. 
They extract concepts as representative training samples or text segments. 
For a fair comparison, we summarize the concept interpretation of these methods using the top 5 representative samples per concept, applying the same summary prompt as ECO-Concept. 
Note that we exclude SelfExplain in global concept comparisons, as it, differently, treats each concept as a single text segment, which cannot be further summarized.

\noindent\textbf{Quantitative Metrics.}
We assess the comprehensibility of concepts using the following metrics:

\noindent\textit{Semantics.~(Sem.)} This metric evaluates the proportion of concepts that have clear semantic meaning. It is calculated as the ratio of concepts with identifiable semantics to the total number of extracted concepts.

\noindent\textit{Distinctiveness.~(Dist.)} This metric evaluates whether the extracted concepts are diverse and not redundant, which is the proportion of unique concepts identified by an LLM in all extracted ones.

\noindent\textit{Consistency.~(Con.)} This metric evaluates the internal consistency of the activation content within each concept. Specifically, we use the topic coherence score $C_V$~\citep{roder2015exploring}, which measures whether the top 10 activated words within each concept collectively form a coherent topic.

As shown in Table~\ref{tab:quantitative_metrics}, unlike post-hoc methods, our method is designed to balance task performance and interpretability simultaneously. However, we still achieve competitive performance in concept comprehensibility across various tasks. In the Hotel task, Cockatiel achieves a higher concept consistency. This is likely because Cockatiel extracts concepts from shorter sentences. However, in other tasks, our method demonstrates comparable or even superior consistency by directly extracting concepts from the original text. For the Twitter task, Concept-Shap achieves a higher proportion of semantically meaningful concepts. Nevertheless, the extracted concepts exhibit relatively low distinctiveness, suggesting that many of the concepts are redundant. In contrast, our method not only extracts semantically meaningful concepts but also ensures they are distinct from each other.
\begin{table*}[ht]
\centering
\caption{Concept comprehensibility evaluation of different concept-based methods. The best result is \colorbox{cell_high}{highlighted}.}
\setlength{\tabcolsep}{1mm}
\resizebox{\textwidth}{!}{
\begin{tabular}{@{}c| ccc ccc ccc ccc ccc ccc ccc@{}}
\toprule
 \multirow{2}{*}{\textbf{Method}} & \multicolumn{3}{c}{\textbf{CEBaB}} & \multicolumn{3}{c}{\textbf{Beer}} & \multicolumn{3}{c}{\textbf{Hotel}} & \multicolumn{3}{c}{\textbf{IMDB}} & \multicolumn{3}{c}{\textbf{AGnews}} & \multicolumn{3}{c}{\textbf{Twitter}} & \multicolumn{3}{c}{\textbf{SciCite}} \\
\cmidrule(lr){2-4} \cmidrule(lr){5-7} \cmidrule(lr){8-10} \cmidrule(lr){11-13} \cmidrule(lr){14-16} \cmidrule(lr){17-19} \cmidrule(lr){20-22}
& Sem. & Dist. & Con. & Sem. & Dist. & Con. & Sem. & Dist. & Con. & Sem. & Dist. & Con. & Sem. & Dist. & Con. & Sem. & Dist. & Con. & Sem. & Dist. & Con. \\
\midrule

Cockatiel & .50 & .40 & .47 & .55 & .60 & .42 & .60 & .65 & \cellcolor{cell_high} \textbf{.49} & .35 & .40 & .41 & .65 & .60 & .41 & .50 & .55 & .53 & .35 & .35 & .50 \\

Concept-Shap & .25 & .30 & .42 & .35 & .15 & .34 & .65 & .40 & .35 & .40 & .30 & .32 & .35 & .35 & .35 & \cellcolor{cell_high} \textbf{.80} & .30 & .53 & .40 & .30 & .44 \\

ProtoTEx & .45 & .45 & .35 & .20 & .25 & .38 & .40 & .30 & .33 & .25 & .25 & .36 & .40 & .45 & .41 & .45 & .50 & .47 & .45 & .45 & .41 \\

ECO-Concept & \cellcolor{cell_high} \textbf{.60} & \cellcolor{cell_high} \textbf{.60} & \cellcolor{cell_high} \textbf{.51} & \cellcolor{cell_high} \textbf{.85} & \cellcolor{cell_high} \textbf{.75} & \cellcolor{cell_high} \textbf{.49} & \cellcolor{cell_high} \textbf{.75} & \cellcolor{cell_high} \textbf{.70} & .48 & \cellcolor{cell_high} \textbf{.65} & \cellcolor{cell_high} \textbf{.65} & \cellcolor{cell_high} \textbf{.52} & \cellcolor{cell_high} \textbf{.70} & \cellcolor{cell_high} \textbf{.65} & \cellcolor{cell_high} \textbf{.54} & .60 & \cellcolor{cell_high} \textbf{.60} & \cellcolor{cell_high} \textbf{.55} & \cellcolor{cell_high} \textbf{.60} & \cellcolor{cell_high} \textbf{.50} & \cellcolor{cell_high} \textbf{.52} \\

\bottomrule
\end{tabular}}
\label{tab:quantitative_metrics}
\vspace{-0.2cm}
\end{table*}

\noindent\textbf{Human Evaluation.}
To further assess whether derived concepts are explainable to humans, we designed three human evaluation tasks, including intruder detection, subjective ratings, and forward simulatability. More details about the survey design and interface are in Appendix~\ref{app:human_evaluation}.

To evaluate the comprehensibility of the extracted concepts, following~\citet{ghorbani2019towards} and~\citet{fel2023craft}, we designed an intruder detection experiment. In each question, participants were presented with four cases and tasked with identifying the one that was conceptually different from the others. A higher accuracy in identifying the intruder indicates that the concepts are more intuitive and easier for humans to understand. 
Table~\ref{tab:intruder_acc} summarizes the results. Our method achieves the highest intruder detection accuracy across all tasks, demonstrating that the concepts extracted by our approach are more comprehensible.
\begin{table}[H]
\centering
\small
\caption{Accuracy of concept intruder detection}
\begin{tabular}{@{}l|ccccccc@{}}
\toprule
\textbf{Method} & \textbf{CEBaB} & \textbf{Beer} & \textbf{Hotel} & \textbf{IMDB} & \textbf{AGnews} & \textbf{Twitter} & \textbf{SciCite} \\ \midrule

Cockatiel & .667 & .633 & .667 & .833 & .767 & .850 & .683 \\
Concept-Shap & .550 & .483 & .483 & .633 & .683 & .850 & .433 \\ 
ProtoTEx & .500 & .500 & .483 & .683 & .750 & .733 & .350 \\
ECO-Concept & \textbf{.767} & \textbf{.750} & \textbf{.767} & \textbf{.900} & \textbf{.900} & \textbf{.867} & \textbf{.700} \\

\bottomrule
\end{tabular}
\label{tab:intruder_acc}
\end{table}

\begin{figure*}[h]
    \centering
    \subfigure[CEBaB]{
    \includegraphics[width=0.23\textwidth]{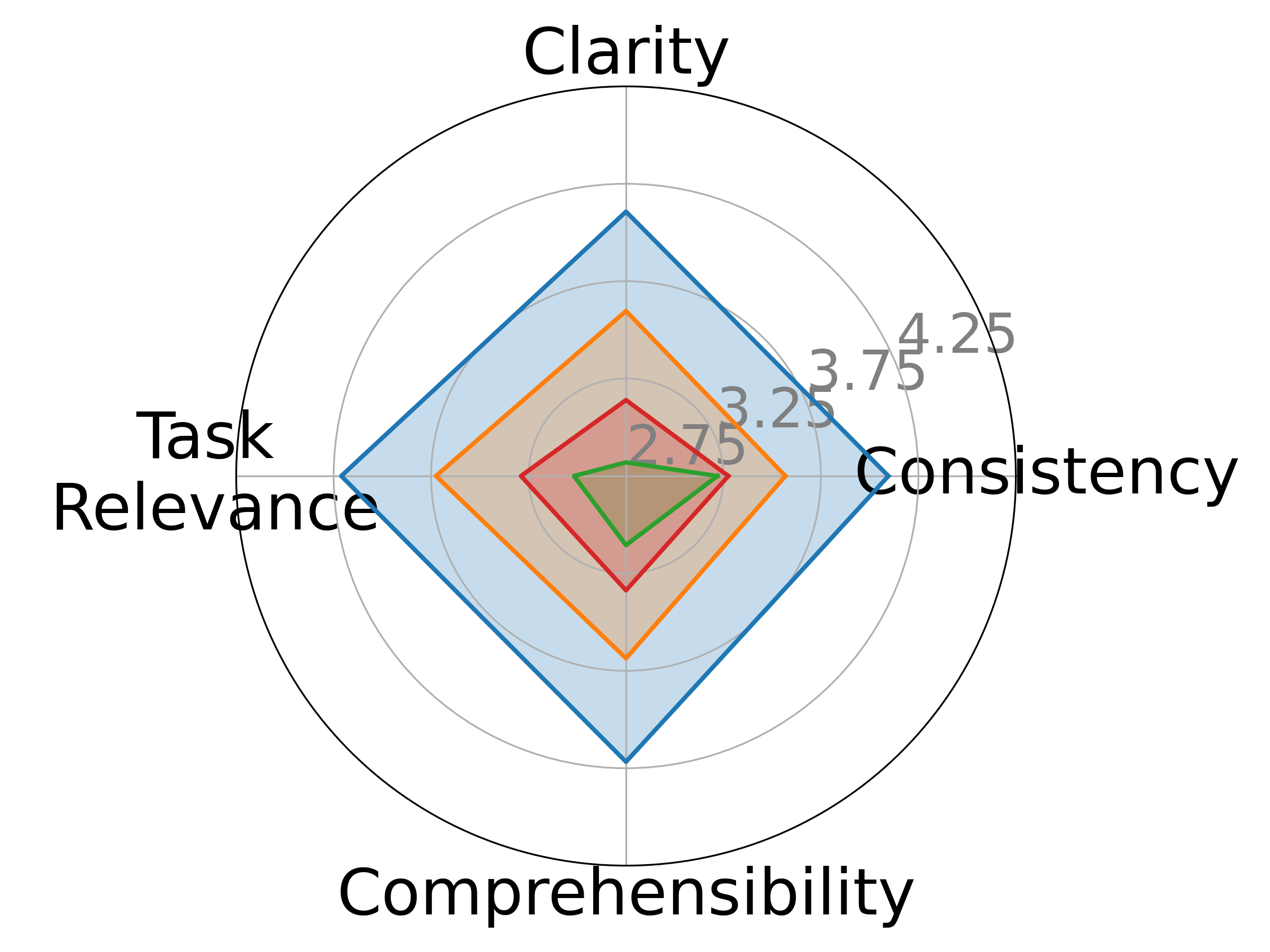}
    }
    \subfigure[Beer]{
    \includegraphics[width=0.23\textwidth]{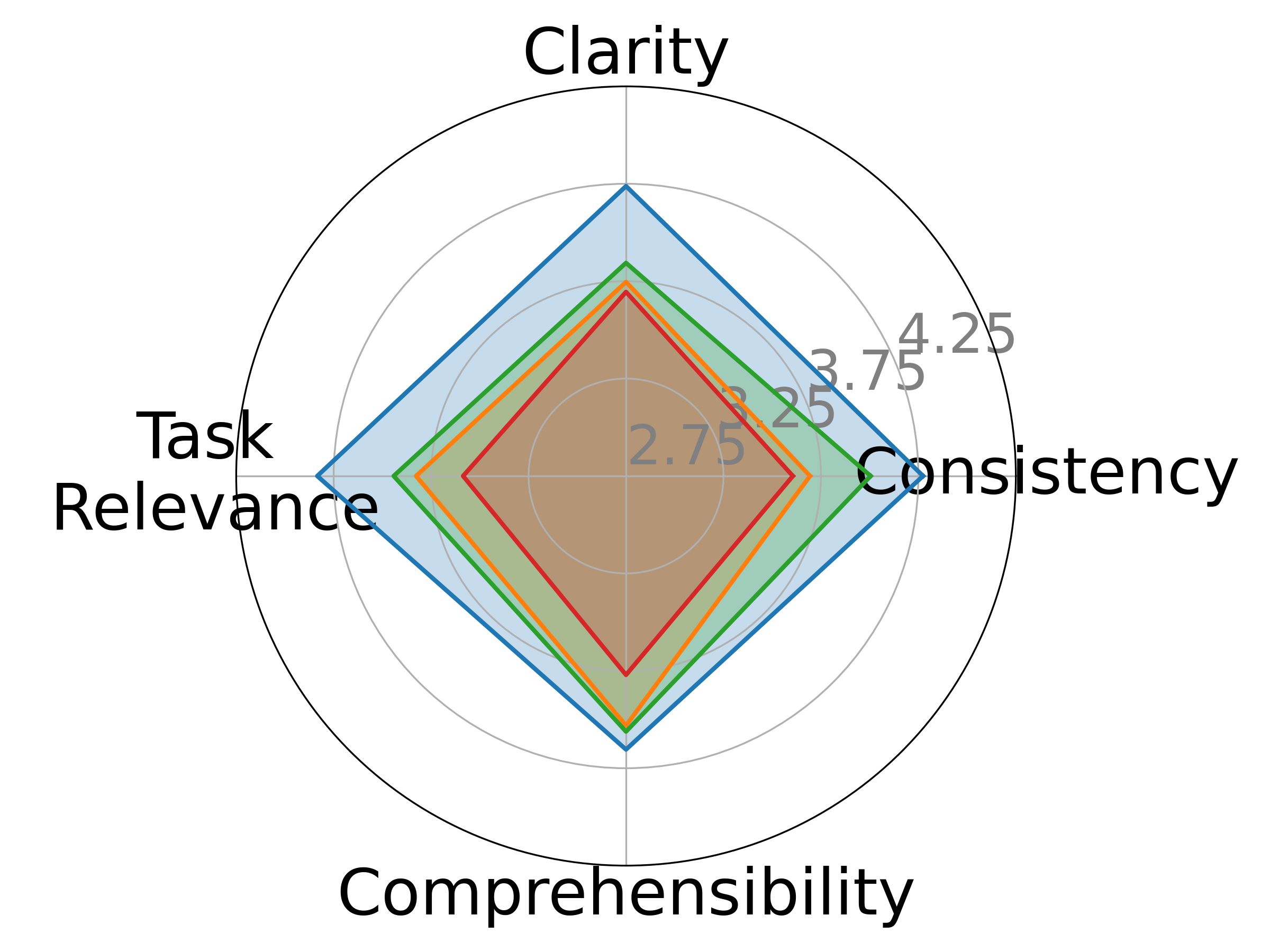}
    }
    \subfigure[Hotel]{
    \includegraphics[width=0.23\textwidth]{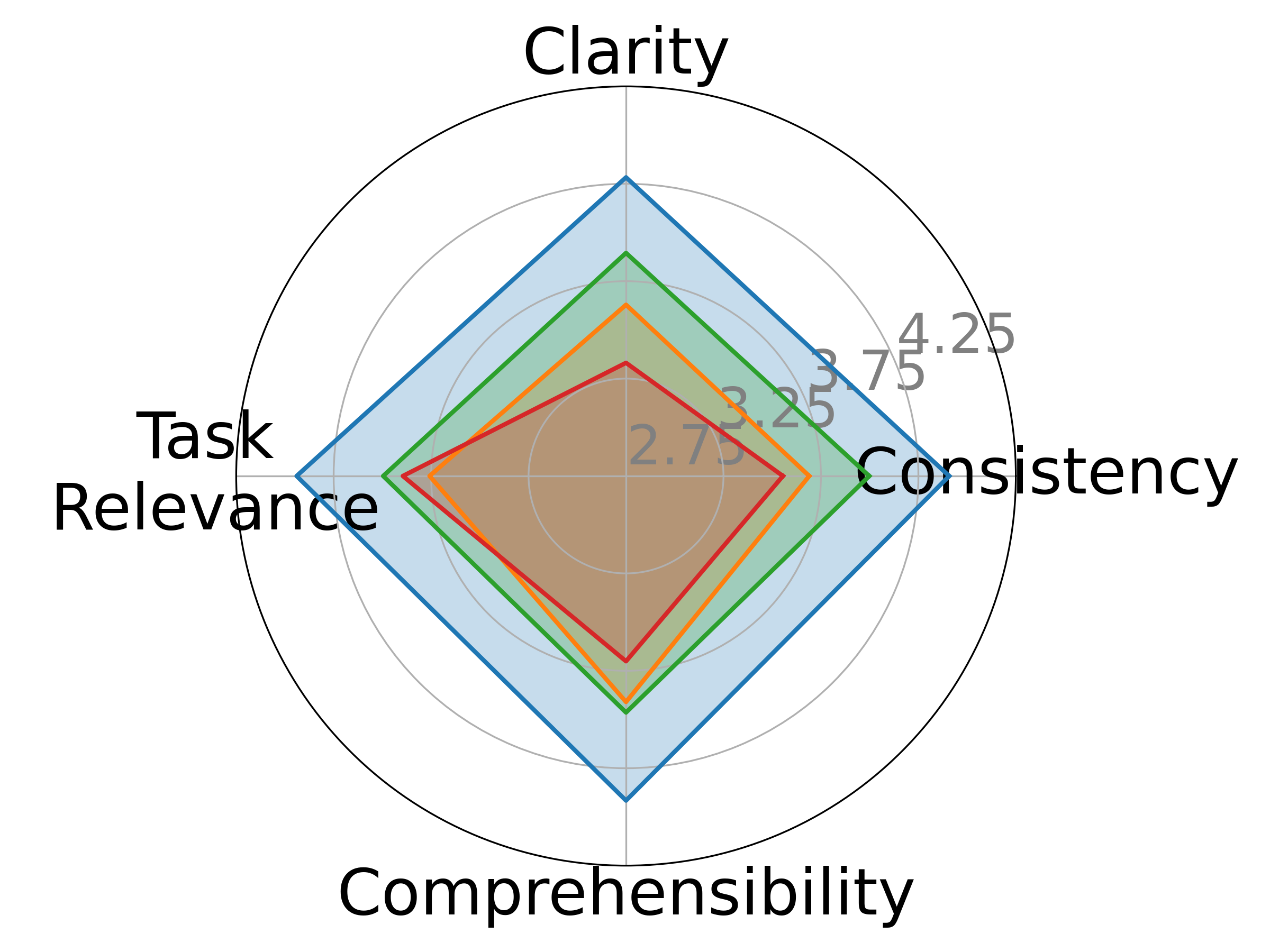}
    }
    \subfigure[IMDB]{
    \includegraphics[width=0.23\textwidth]{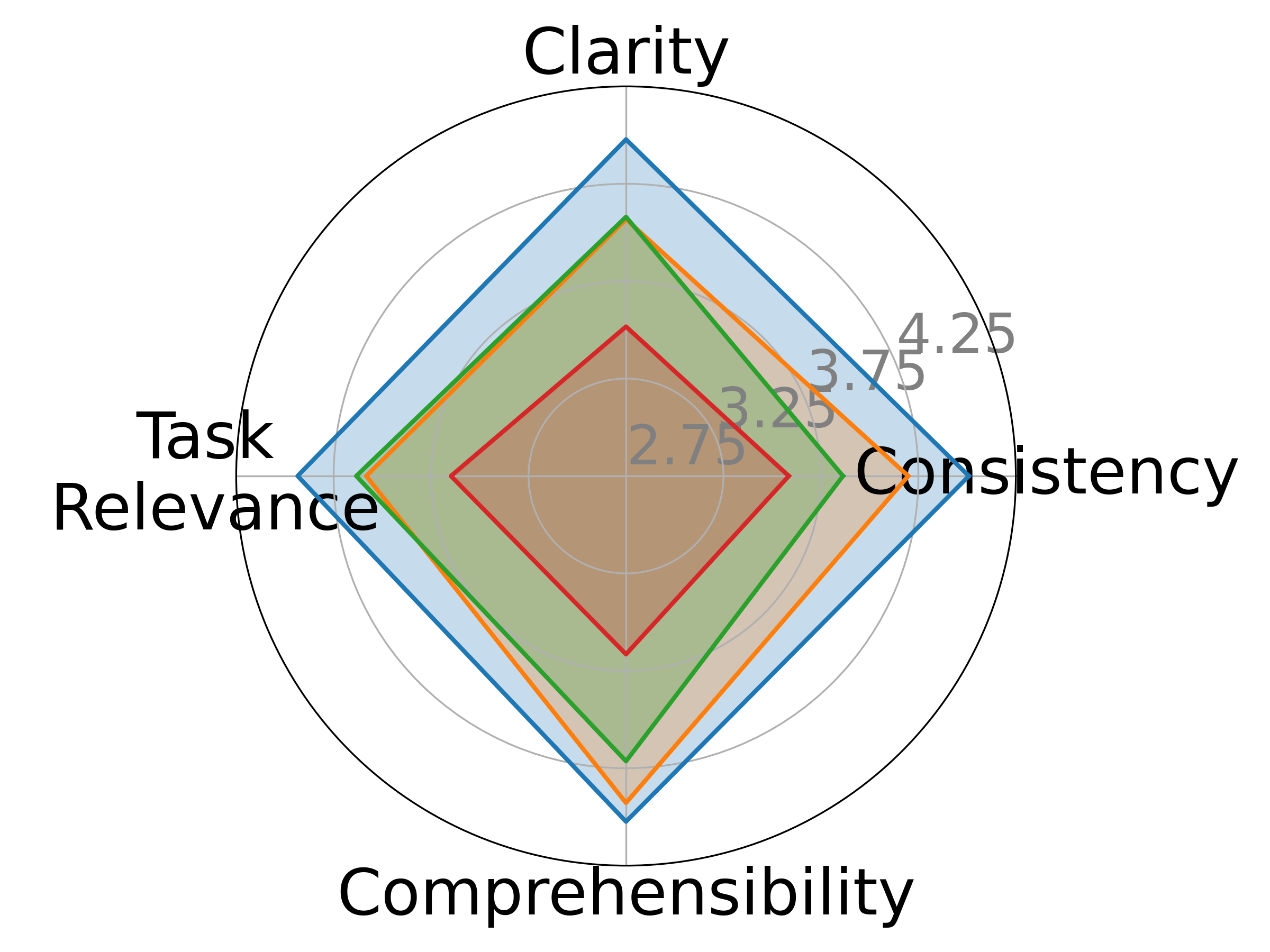}
    }
    \subfigure[AGnews]{
    \includegraphics[width=0.23\textwidth]{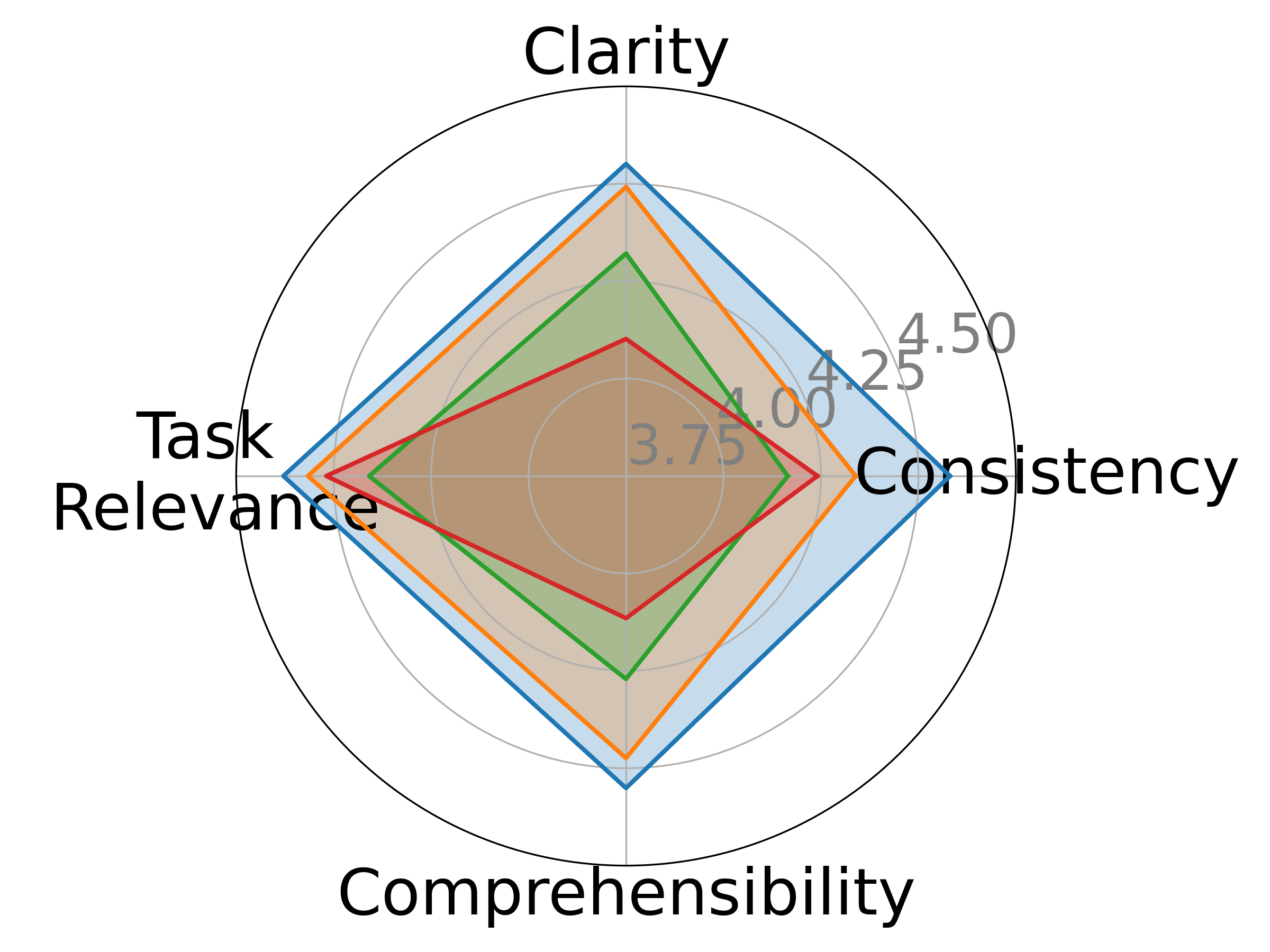}
    }
    \subfigure[Twitter]{
    \includegraphics[width=0.23\textwidth]{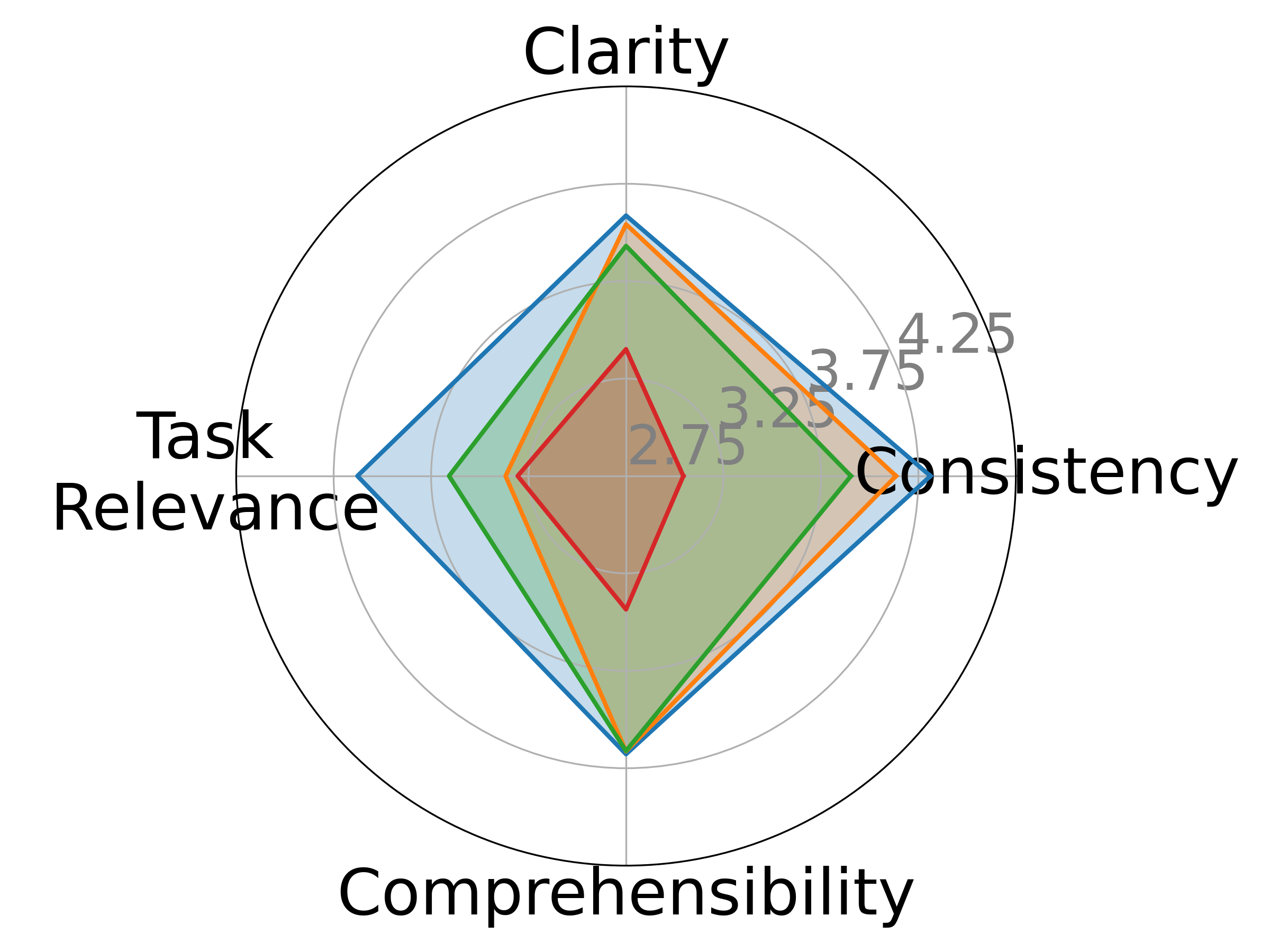}
    }
    \subfigure[SciCite]{
    \includegraphics[width=0.23\textwidth]{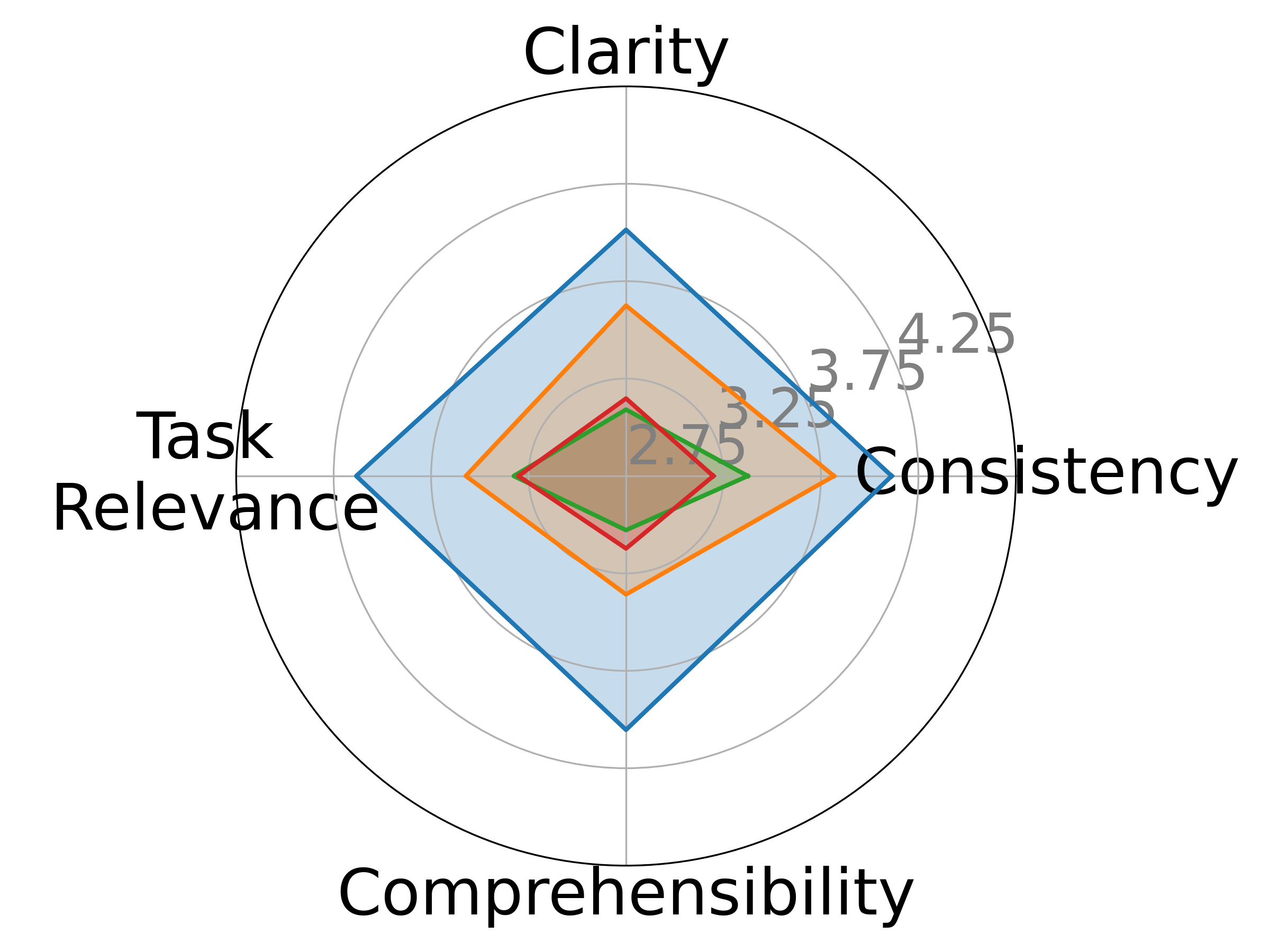}
    }
    \subfigure[Legend]{
    \includegraphics[width=0.23\textwidth]
    {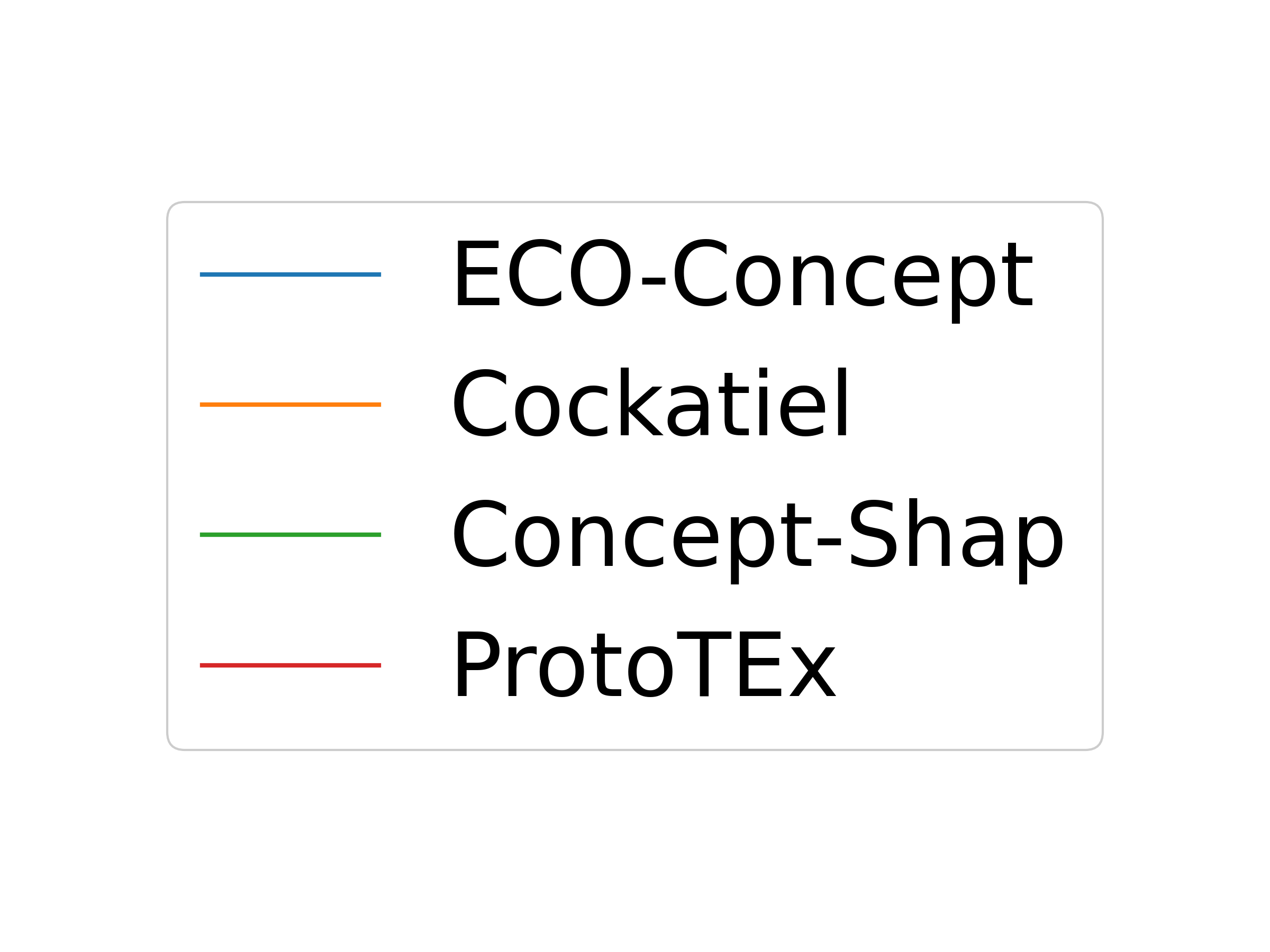}
    }
    \caption{Human subjective ratings on concept quality}
    \label{figs:human_ratings}
\end{figure*}

In addition, we also conducted subjective rating experiments, where participants rated all the extracted concepts from multiple perspectives: \textit{Consistency, Clarity, Task Relevance, Comprehensibility}. As shown in Fig.~\ref{figs:human_ratings}, our method consistently achieved the highest ratings across all perspectives for each task. This result indicates that our concepts are subjectively the most comprehensible to humans and are closely aligned with the tasks. In contrast, ProtoTEx received the lowest ratings in most tasks, suggesting that traditional unsupervised self-explaining methods often struggle to balance interpretability with task performance, resulting in less intuitive concepts for human understanding.

Finally, we validated the extracted concepts through a forward simulatability experiment to evaluate their effectiveness in helping users understand the model's behavior. This aligns with the definition of explainability proposed by~\citet{kim2016examples}, which emphasizes a user's ability to "correctly and efficiently predict the method’s results." 
In our experiment, participants received explanations of model outputs, then attempted to infer the model's outputs based on the explanations and rated their confidence.
For comparison, we also measured the accuracy of human judgments without explanations~(NE). 
We conducted forward simulatability experiments with explanations on the two most challenging tasks for human judgment~(Beer and AGnews), as determined by the lowest human accuracy rates in no-explanation conditions.
As shown in Table~\ref{tab:explanation_acc}, our method significantly improved the simulatability accuracy for participants in inferring model outputs. This demonstrates that our explanations effectively enhance human understanding of the model's behavior and increase their trust in its predictions.
Moreover, for other concept-based methods, although the simulatability accuracy is higher when their explanations are provided compared to when no explanations are given, humans generally tend to have lower confidence in these explanations.
This further validates that existing methods often produce confusing explanations, which in turn reduces human trust.
Besides, we also asked participants to rate the explanations for each example across three aspects. Detailed information and results are in Appendix~\ref{app:forward_simu}.

\begin{table}[H]
\centering
\small
\caption{Results of human forward simulatability}
\begin{tabular}{@{}l|cc cc@{}}
\toprule
\multirow{2}{*}{\textbf{Method}} & \multicolumn{2}{c}{\textbf{Beer}} & \multicolumn{2}{c}{\textbf{AGnews}} \\
\cmidrule(lr){2-3} \cmidrule(lr){4-5} 
 & Accuracy & Confidence & Accuracy & Confidence \\
\midrule
NE & .808 & 4.52 & .708 & 4.44  \\
Cockatiel & .967 & 4.13 & .833 & 4.50  \\
Concept-Shap & .833 & 4.22 & .783 & 3.80  \\
PROTOTEX & .950 & 4.28 & .783 & 4.23 \\
ECO-Concept & \textbf{.983} & \textbf{4.45} & \textbf{.867} & \textbf{4.55}  \\
\bottomrule
\end{tabular}
\label{tab:explanation_acc}
\end{table}

\subsection{Ablation Study}
To further explore the impact of different model modules, we conducted several ablation studies. 

\noindent\textbf{Impact of Concept Regularizers.}
To analyze the effects of the concept consistency and concept distinctiveness regularizers, we conduct an ablation study. 
We experiment on two types of variant models: w/o $\mathcal{L}_{con}$ and w/o $\mathcal{L}_{dist}$, which respectively remove the concept consistency regularizer and the concept distinctiveness regularizer. Experimental results are shown in Table~\ref{tab:impact_of_r}.
The results show that the best task performance and concept interpretability are achieved when both regularizers are applied together. 
By comparing the w/ all and w/o $\mathcal{L}_{con}$ variants, removing the consistency loss significantly reduces the consistency score, demonstrating the importance of the consistency regularizer. A similar trend is observed with the distinctiveness regularizer.
These results highlight the importance of incorporating both types of regularization. 
Besides, we observed that the decrease in the \textit{Distinctiveness} score when removing $\mathcal{L}_{con}$ (from w/ all to w/o $\mathcal{L}_{con}$ variant) is greater than the decrease when removing $\mathcal{L}_{dist}$ (from w/ all to w/o $\mathcal{L}_{dist}$ variant). This occurs in the tasks (IMDB, AGnews, Twitter) where the \textit{Semantics} score of w/o $\mathcal{L}_{con}$ variant is consistently lower than that of w/o $\mathcal{L}_{dist}$ variant. This indicates that removing the consistency regularizer in these tasks leads to fewer discovered concepts with clear semantic meaning, which in turn affects the calculation of concept distinctiveness. This further supports the rationale for using both regularizers simultaneously. The two regularizers mutually reinforce each other, and their combined use is crucial for discovering more comprehensible concepts.
\begin{table}[H]
\centering
\small
\caption{Task performance and concept metrics comparison between our method and its ablative variants}
\resizebox{0.65\textwidth}{!}{
\begin{tabular}{@{}l|l|ccccccc@{}}
\toprule
\textbf{Metric} & \textbf{Method} & \textbf{CEBaB} & \textbf{Beer} & \textbf{Hotel} & \textbf{IMDB} & \textbf{AGnews} & \textbf{Twitter} & \textbf{SciCite} \\ \midrule

\multirow{4}{*}{Acc} 
& w/o $\mathcal{L}_{con}$ & .679 & \cellcolor{cell_high} \textbf{.885} & .980 & .937 & .939 & .823 & .858 \\  
& w/o $\mathcal{L}_{dist}$ & .699 & \cellcolor{cell_high} \textbf{.885} & .977 & .937 & .940 & .824 & .852 \\  
& w/ all & \cellcolor{cell_high} \textbf{.704} & \cellcolor{cell_high} \textbf{.885} & \cellcolor{cell_high} \textbf{.981} & \cellcolor{cell_high} \textbf{.938} & \cellcolor{cell_high} \textbf{.945} &  \cellcolor{cell_high} \textbf{.832} & \cellcolor{cell_high} \textbf{.860} \\  
\midrule

\multirow{4}{*}{F1} 
& w/o $\mathcal{L}_{con}$ & .796 & \cellcolor{cell_high} \textbf{.885} & .980 & .937 & .959 & .811 & .879 \\  
& w/o $\mathcal{L}_{dist}$ & .806 & \cellcolor{cell_high} \textbf{.885} & .977 & .937 & .960 & .809 & .875 \\  
& w/ all & \cellcolor{cell_high} \textbf{.811} & \cellcolor{cell_high} \textbf{.885} & \cellcolor{cell_high} \textbf{.981} & \cellcolor{cell_high} \textbf{.938} & \cellcolor{cell_high} \textbf{.964} & \cellcolor{cell_high} \textbf{.816} & \cellcolor{cell_high} \textbf{.882} \\  
\midrule

\multirow{4}{*}{Sem.} 
& w/o $\mathcal{L}_{con}$ & .45 & .65 & \cellcolor{cell_high} \textbf{.70} & .40 & .50 & .40 & .25 \\
& w/o $\mathcal{L}_{dist}$ & .50 & \cellcolor{cell_high} \textbf{.70} & \cellcolor{cell_high} \textbf{.70} & .50 & \cellcolor{cell_high} \textbf{.65} & .50 & .15 \\ 
& w/ all & \cellcolor{cell_high} \textbf{.55} & \cellcolor{cell_high} \textbf{.70} & \cellcolor{cell_high} \textbf{.70} & \cellcolor{cell_high} \textbf{.60} & \cellcolor{cell_high} \textbf{.65} & \cellcolor{cell_high} \textbf{.60} & \cellcolor{cell_high} \textbf{.55} \\
\midrule

\multirow{4}{*}{Dist.} 
& w/o $\mathcal{L}_{con}$ & .50 & \cellcolor{cell_high} \textbf{.65} & .60 & .40 & .45 & .50 & .35 \\
& w/o $\mathcal{L}_{dist}$ & .30 & .50 & .55 & .50 & .55 & .55 & .20 \\ 
& w/ all & \cellcolor{cell_high} \textbf{.60} & \cellcolor{cell_high} \textbf{.65} & \cellcolor{cell_high} \textbf{.65} & \cellcolor{cell_high} \textbf{.60} & \cellcolor{cell_high} \textbf{.65} & \cellcolor{cell_high} \textbf{.60} & \cellcolor{cell_high} \textbf{.50} \\
\midrule

\multirow{4}{*}{Con.} 
& w/o $\mathcal{L}_{con}$ & .43 & .40 & .41 & .38 & .47 & .45 & .42 \\
& w/o $\mathcal{L}_{dist}$ & \cellcolor{cell_high} \textbf{.48} & .41 & .41 & .40 & .51 & .45 & .42 \\
& w/ all & \cellcolor{cell_high} \textbf{.48} & \cellcolor{cell_high} \textbf{.46} & \cellcolor{cell_high} \textbf{.43} & \cellcolor{cell_high} \textbf{.49} & \cellcolor{cell_high} \textbf{.52} & \cellcolor{cell_high} \textbf{.51} & \cellcolor{cell_high} \textbf{.47} \\
\bottomrule
\end{tabular}}
\label{tab:impact_of_r}
\end{table}

\begin{figure*}[h]
    \centering
    \subfigure[CEBaB]{
    \includegraphics[width=0.23\textwidth]{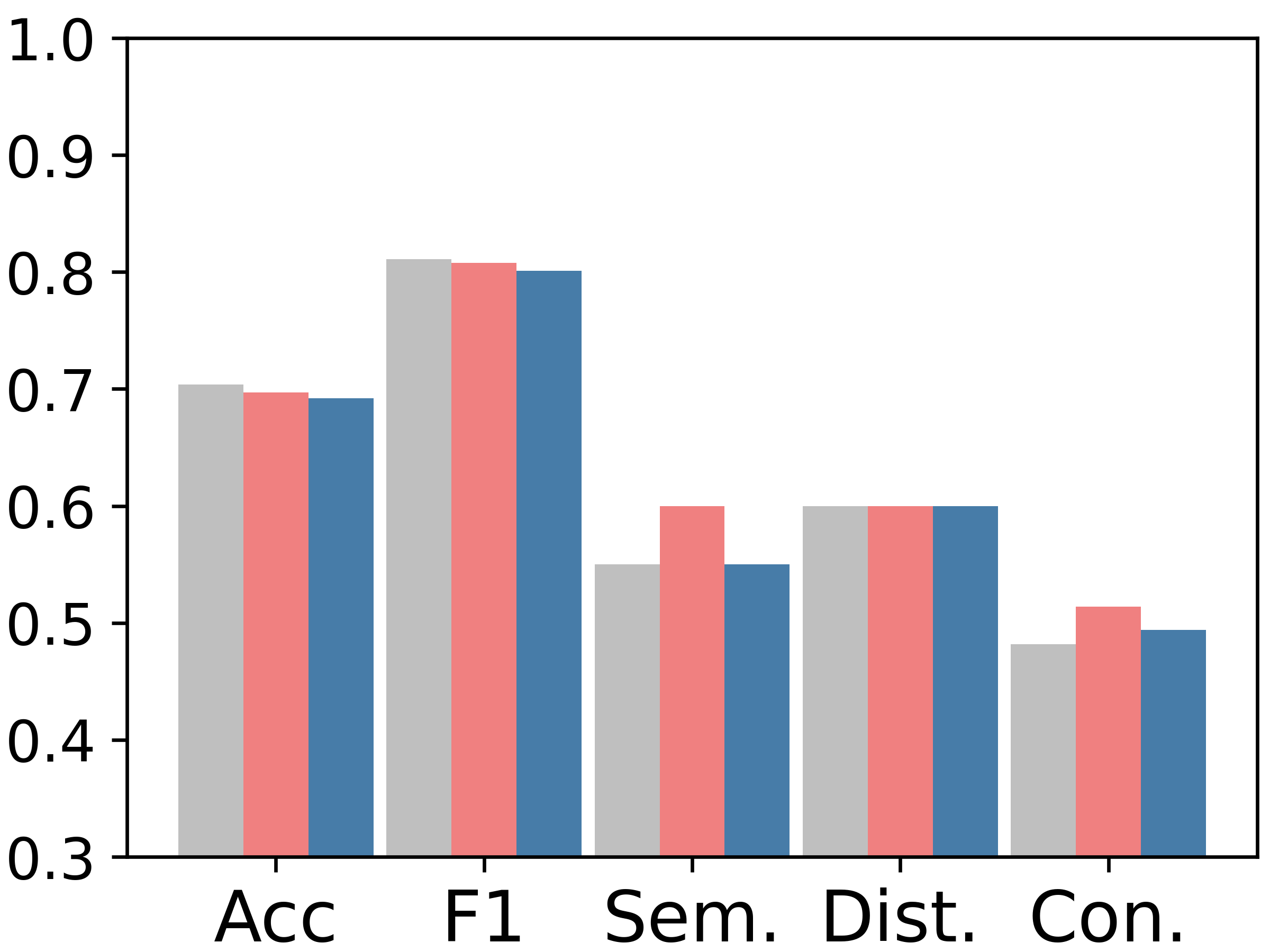}
    }
    \subfigure[Beer]{
    \includegraphics[width=0.23\textwidth]{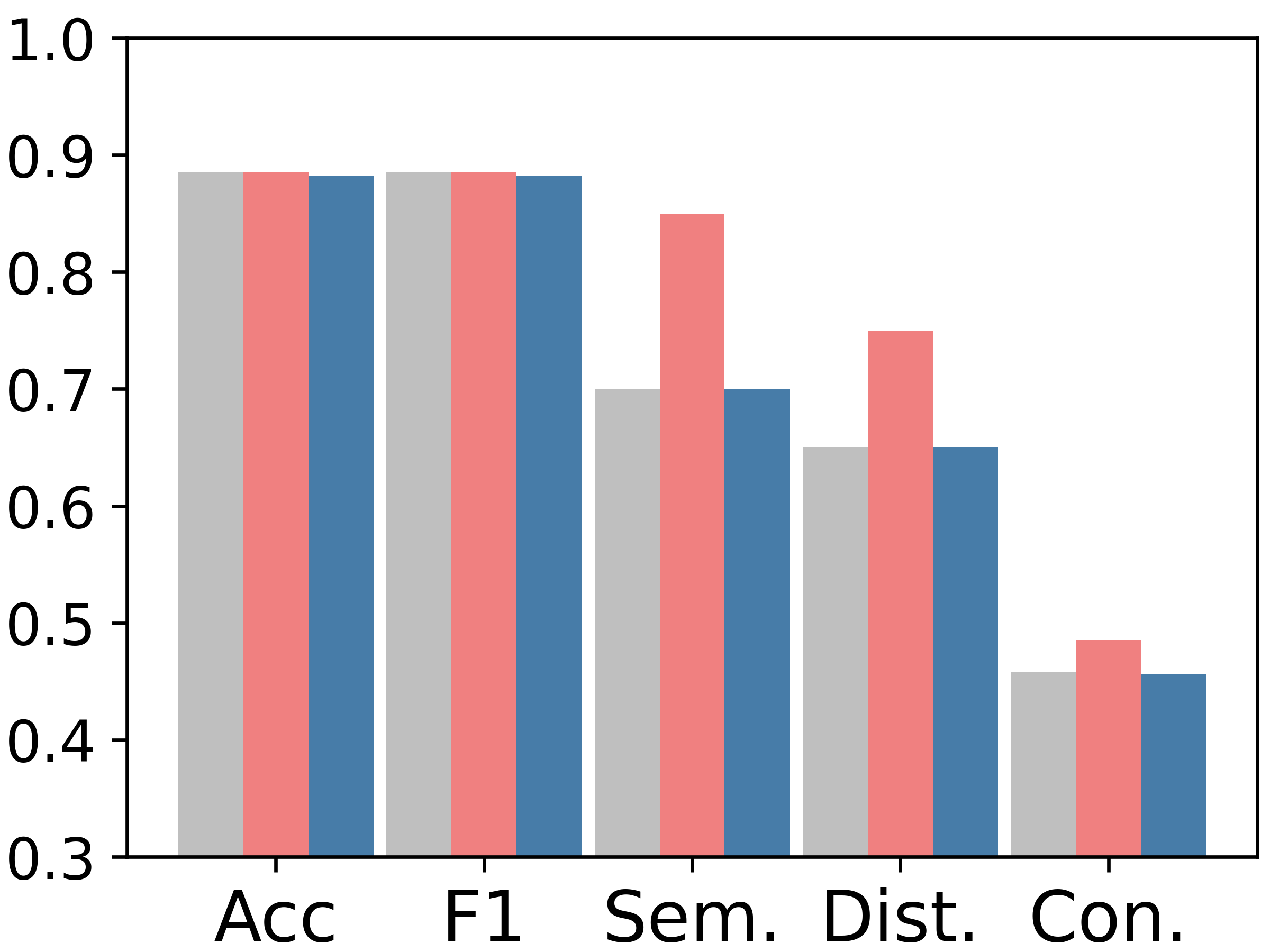}
    }
    \subfigure[Hotel]{
    \includegraphics[width=0.23\textwidth]{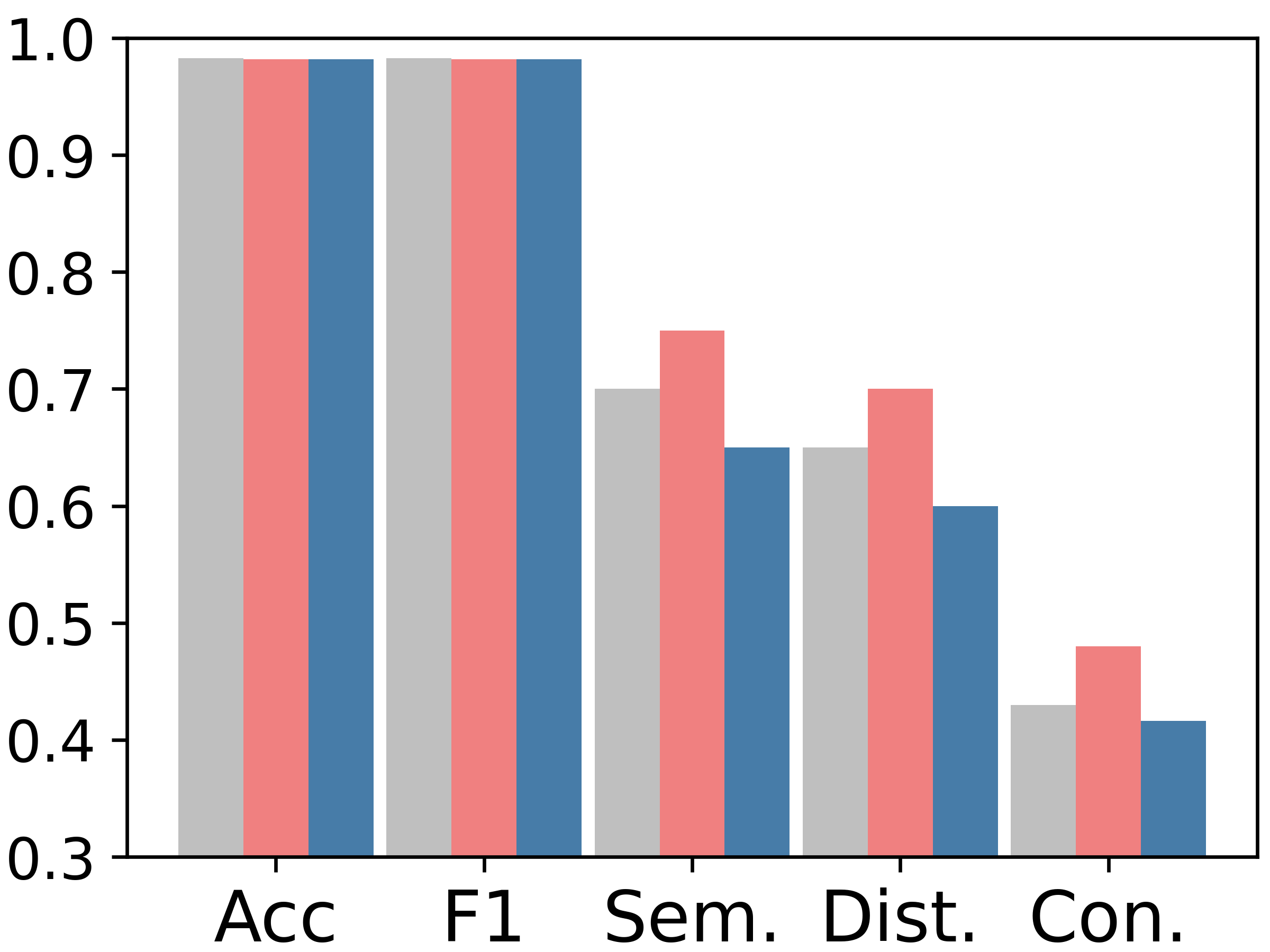}
    }
    \subfigure[IMDB]{
    \includegraphics[width=0.23\textwidth]{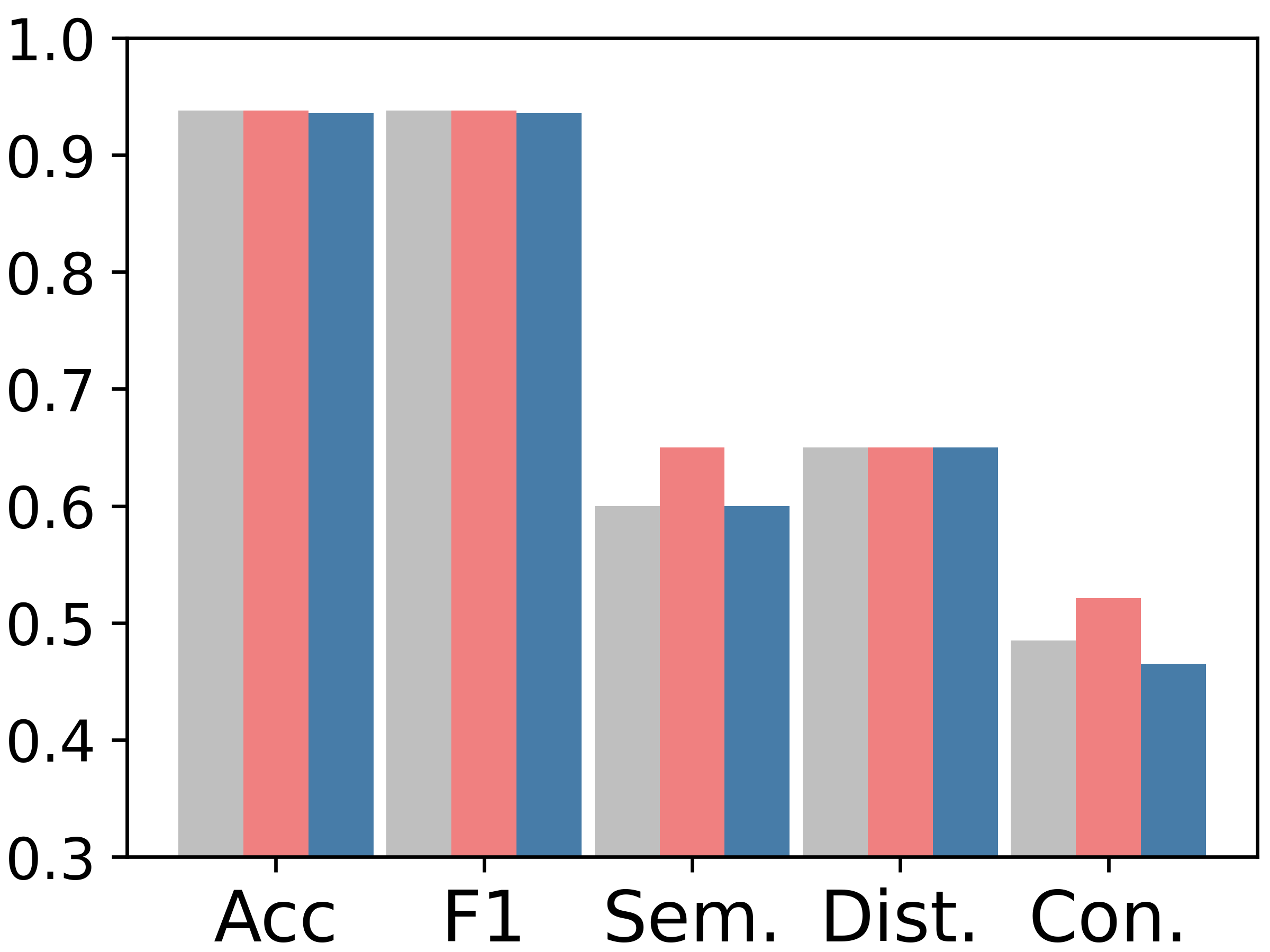}
    }
    \subfigure[AGnews]{
    \includegraphics[width=0.23\textwidth]{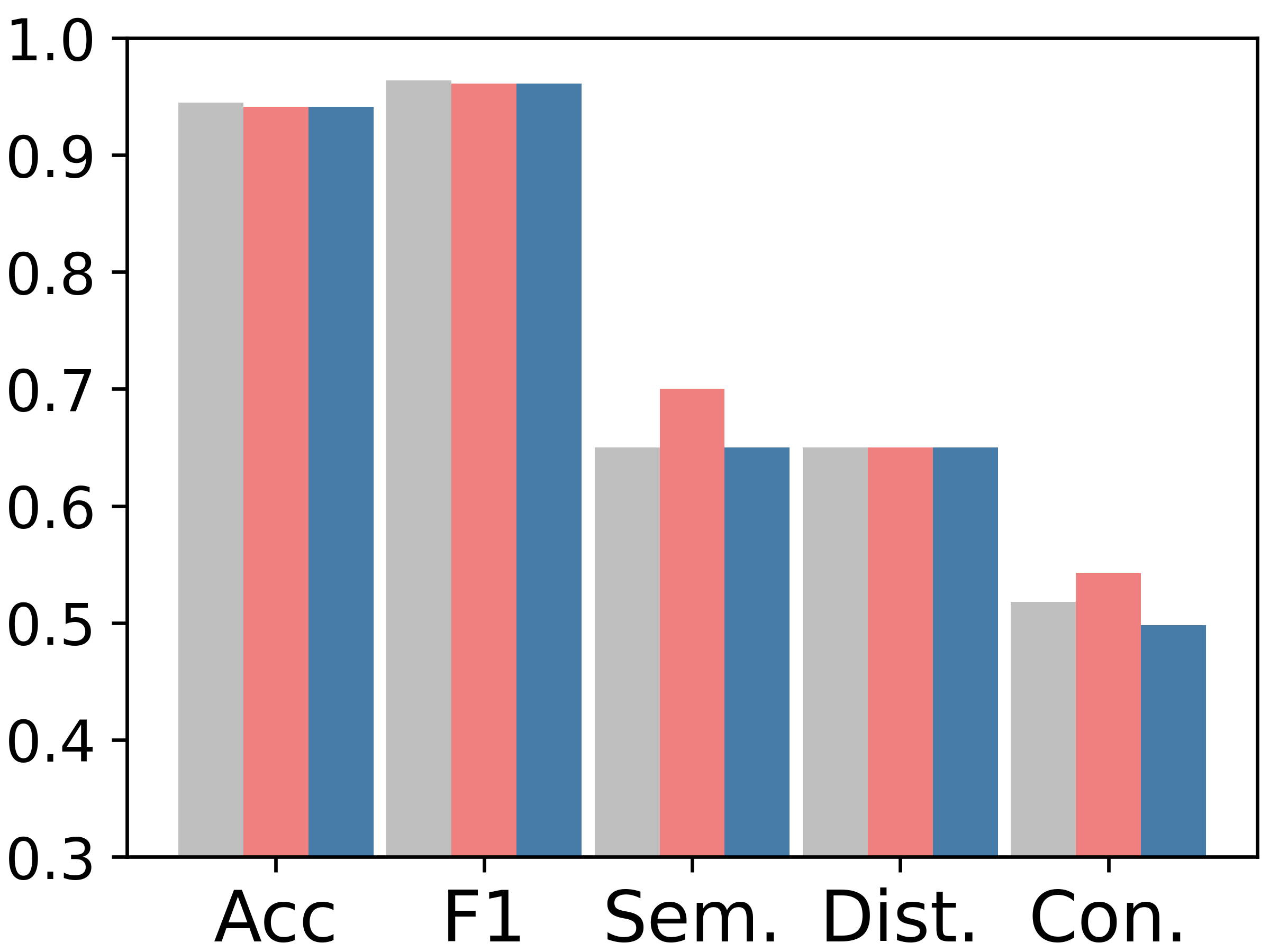}
    }
    \subfigure[Twitter]{
    \includegraphics[width=0.23\textwidth]{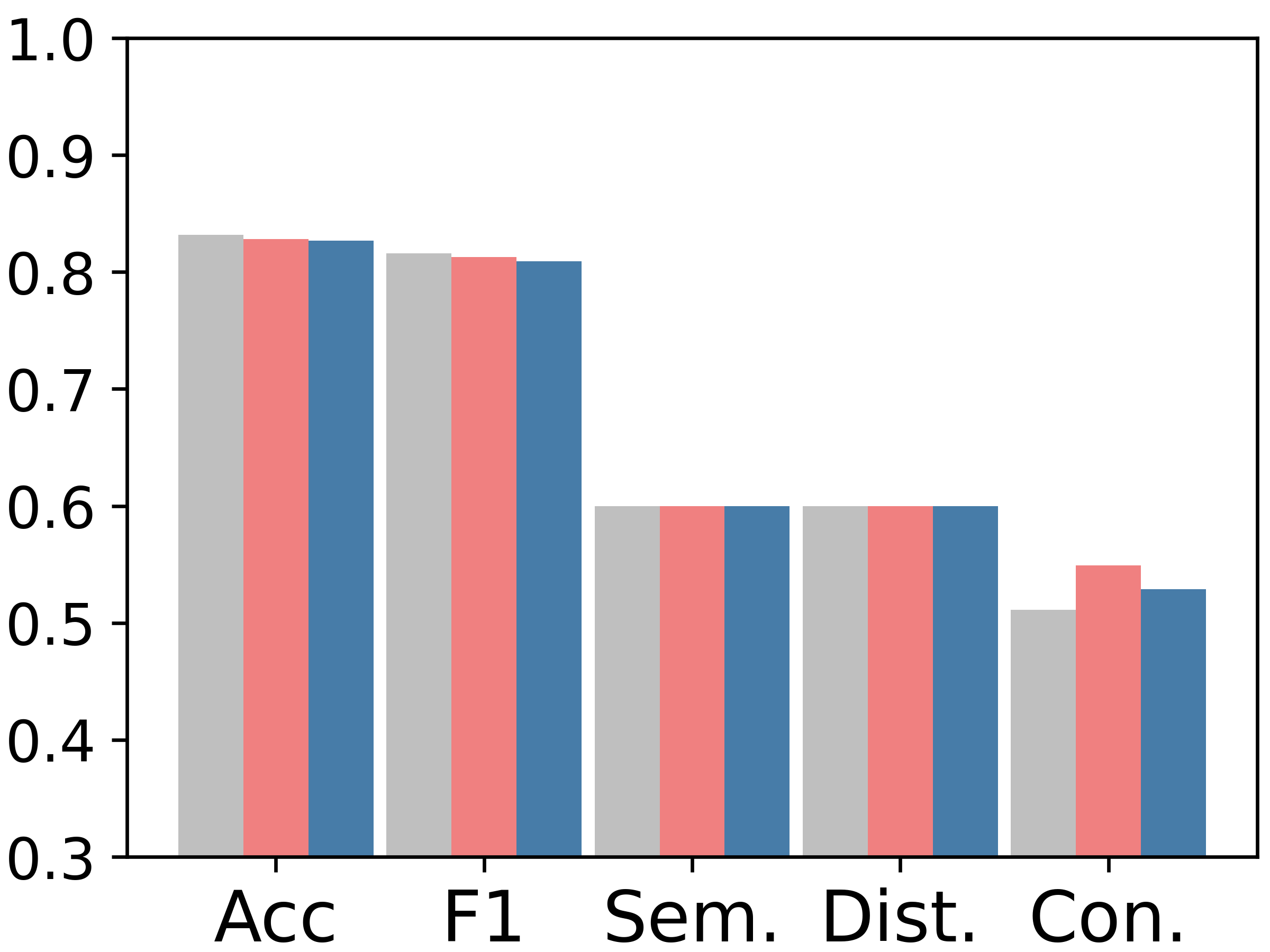}
    }
    \subfigure[SciCite]{
    \includegraphics[width=0.23\textwidth]{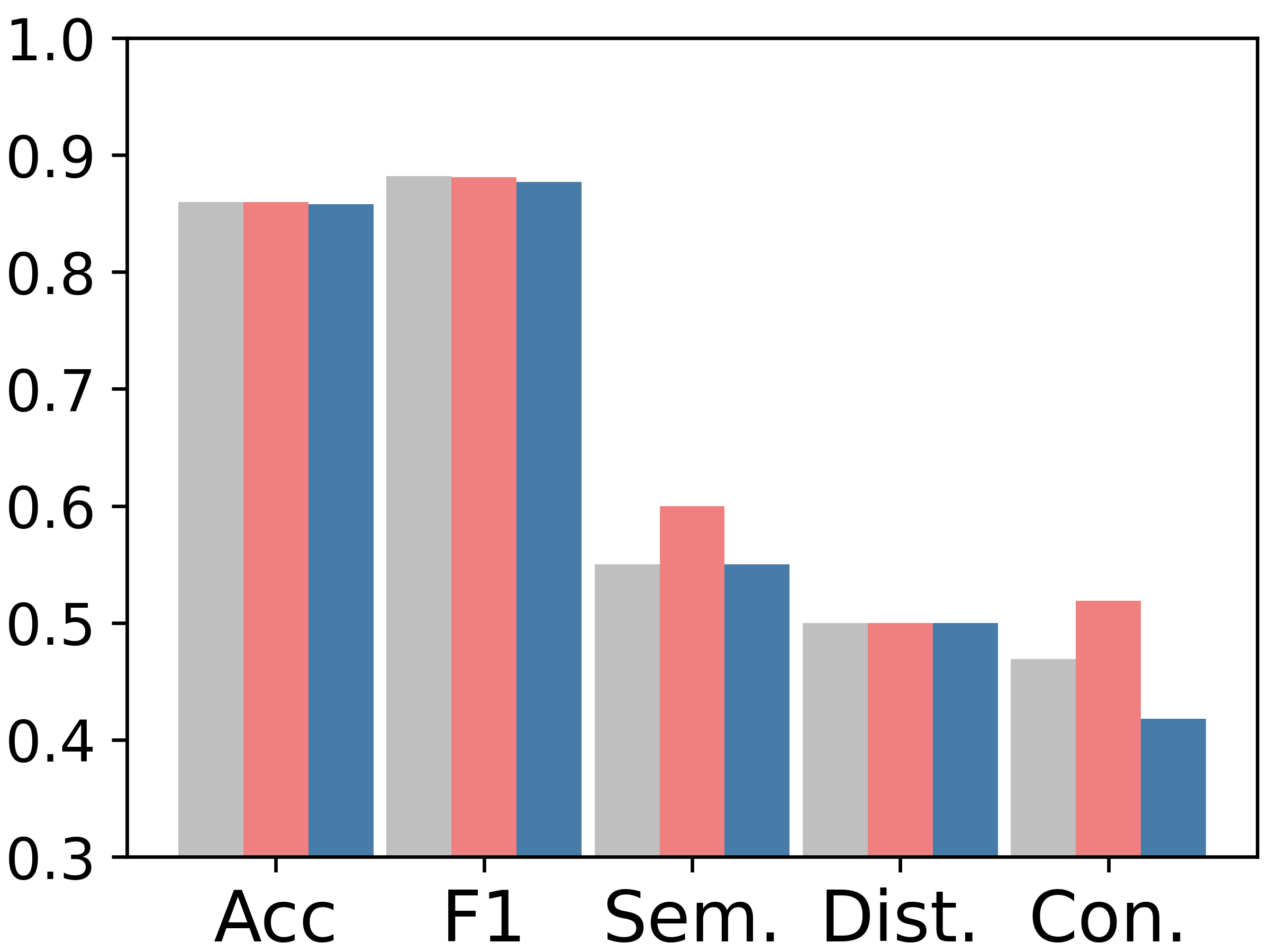}
    }
    \subfigure[Legend]{
    \includegraphics[width=0.23\textwidth]
    {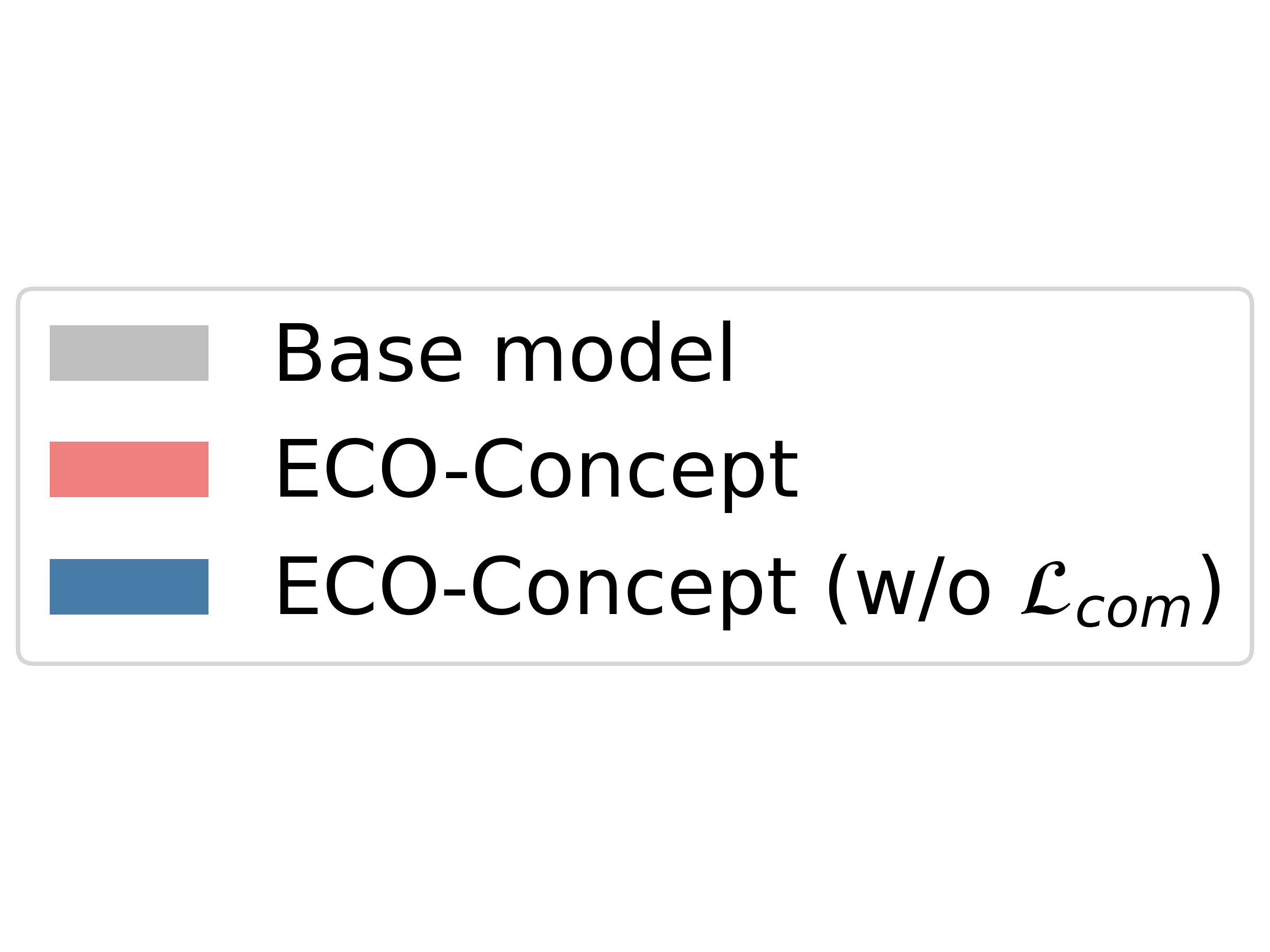}
    }
    \caption{Classification performance and concept metrics comparison of the model before concept enhancement~(the Base model), ECO-Concept, and ECO-Concept~(w/o $\mathcal{L}_{com}$).}
    \label{figs:impact_of_e}
\end{figure*}

\noindent\textbf{Impact of Concept Comprehensibility Enhancement.}
To demonstrate the importance of the concept comprehensibility enhancement stage, we compare the model before enhancement~(denoted as the Base model), ECO-Concept, and ECO-Concept~(w/o $\mathcal{L}_{com}$). We introduce the ECO-Concept~(w/o $\mathcal{L}_{com}$) variant to exclude the effect of longer training with concept regularizers, ensuring that any improvements in comprehensibility are due to $\mathcal{L}_{com}$. Both ECO-Concept and ECO-Concept (w/o $\mathcal{L}_{com}$) are fine-tuned from the Base model. The ECO-Concept~(w/o $\mathcal{L}_{com}$) variant is fine-tuned using the same parameter settings and number of iterations as ECO-Concept training, but without the comprehensibility loss.
From the result in Fig.~\ref{figs:impact_of_e}, we observe that after concept enhancement, classification performance remains largely unchanged (or experiences only a slight decline). 
Regarding concept comprehensibility metrics, our ECO-Concept yields the best performance.
After the enhancement, the semantics and consistency of concepts are further improved across most tasks. Besides, we also discovered more diverse concepts in two tasks. Further analysis is provided in Appendix~\ref{app:impact_of_e}. 

\section{Conclusion}
\label{conclusion}
To automatically extract human-understandable concept explanations with no predefined concept annotations, we proposed ECO-Concept, an intrinsically interpretable framework. 
ECO-Concept employs an object-centric architecture based on the slot attention mechanism to extract task-specific semantic concepts. To ensure the extracted concepts are comprehensible to humans, ECO-Concept incorporates LLMs as human proxies to evaluate concept comprehensibility during model training, using their feedback to iteratively refine the concept extractor.
ECO-Concept achieves classification performance on par with black-box models while offering enhanced interpretability. It outperforms existing concept-based approaches in both quantitative metrics and user studies, demonstrating its effectiveness in generating human-aligned, interpretable explanations.

\section*{Limitations}
While our approach presents a significant step towards more interpretable models, several limitations warrant further exploration. First, in this work, we focus on enhancing pre-trained language models with self-explainable structures to improve interpretability in text classification tasks. Given the inherent challenges of unsupervised concept extraction, we employ a comparatively lightweight BERT-based model as the backbone. For future work, we plan to explore concept extraction methods compatible with larger models and extend our framework to encompass state-of-the-art architectures such as LLaMA and Mixtral. Second, to ensure the assessment of concept comprehensibility is close to the human level, we utilized two well-recognized API-based LLMs based on their strong instruction-following capabilities. However, due to cost limitations, we were only able to simulate concepts on a subset of samples. The current setting represents the best trade-off between performance and cost within our acceptable range. In the future, we plan to test more cost-effective and deployable open-source LLMs. Lastly, in our method, the number of concepts is fixed and cannot be adaptively adjusted during training. Currently, there are no well-suited approaches to address this challenge. We are actively exploring potential techniques~(such as incremental learning) to adjust the number of concepts in a more flexible manner.

\bibliography{main}

\clearpage

\appendix
\section{Experimental Details}
\label{app:experimental_details}
\subsection{Datasets}
\label{app:datasets}
In this section, we provide detailed descriptions of the benchmark datasets used in our experiments.

\textbf{CEBaB}~\citep{abraham2022cebab} is a commonly used dataset for concept-based text analysis. It includes restaurant reviews,
annotated with sentiment ratings (ranging from 1 to 5 stars) and four dining experience concepts: food quality, noise level, ambiance, and service. Following~\citet{tan2024sparsity}, we frame this as a five-class classification task.

\textbf{Beer}~\citep{mcauley2012learning} is a multi-aspect beer reviews dataset. Each review includes sentiment ratings across five aspects: appearance, aroma, palate, taste, and overall impression. Following~\citet{jourdan2023cockatiel}, we train the model to predict whether the overall score exceeds 3, indicating a positive review. The remaining four aspects are treated as concept labels for supervised concept-based methods.

\textbf{Hotel}~\citep{wang2010latent} is a multi-aspect hotel reviews dataset, comprising review texts annotated with seven concept labels: value, rooms, location, cleanliness, check-in/front desk, service, and business service. Since the original labels are on a scale of 0 to 5, we utilize the binarized version proposed by~\citet{bao2018deriving}. 

\textbf{IMDB}~\citep{maas2011learning} is a movie reviews dataset, where each review is labeled with either positive or negative sentiment.

\textbf{AGnews}~\citep{gulli2004ag} is a dataset of news articles categorized into one of four classes: business, science/technology, sports, or world/political.

\textbf{Twitter}~\citep{sheng2021integrating} is a dataset consisting of real and fake news collected from Twitter.

\textbf{SciCite}~\citep{cohan2019structural} is a dataset designed for classifying citation intents in academic papers. Each citation is categorized into one of three classes: method, background, or result.

\subsection{Baselines}
\label{app:baselines}
The details of the baseline methods are shown as follows.

\textbf{BERT-/RoBERTa-based classifier}~\citep{devlin2018bert,liu2019roberta}: We utilize BERT/RoBERTa to encode tokens of text content and feed the extracted average embedding into an MLP to obtain the final prediction.

\textbf{CBM}~\citep{kim2018interpretability} introduced an intermediate layer that first predicts the human-defined concepts and then uses the concepts to predict the final output. 

\textbf{SparseCBM}~\citep{tan2024sparsity} enhances CBM by integrating unstructured pruning. SparseCBM constructs concept-specific sparse subnetworks within the backbone network, providing interpretability while retaining model performance. 

\textbf{SelfExplain}~\citep{rajagopal2021selfexplain} is an unsupervised self-explaining model. It provides both global and local concept explanations for each sample while performing the classification tasks. 

\textbf{PROTOTEX}~\citep{das2022prototex} is an unsupervised self-explaining classification architecture based on prototype networks. It explains model decisions based on prototype tensors that encode latent clusters of training examples.

\textbf{Cockatiel}~\citep{jourdan2023cockatiel} is an unsupervised post-hoc concept-based method. It generates meaningful concepts from the last layer of a neural net model trained on an NLP classification task by using Non-Negative Matrix Factorization. In our experiment, we use a RoBERTa-based classifier as the base model for concept extraction.

\textbf{Concept-Shap}~\citep{yeh2020completeness} is an unsupervised post-hoc concept-based method that aims to infer a complete set of concepts. In our experiment, we use a RoBERTa-based classifier as the base model for concept extraction.

It is worth noting that we did not include approaches that utilize LLMs to extract concepts. These methods generate concept sets using LLMs before training and follow a pipeline similar to that of CBM. Specifically, in the text domain, their primary distinction from CBM lies in the use of concept sets defined by LLMs or humans. As such, we use CBM as a representative method.

\subsection{Computational Complexity Analysis}
Compared to the baseline methods, incorporating LLM-based evaluation during training increases computational costs. Specifically, our method requires approximately 22.2\% more resources than the best-performing baseline, Cockatiel. However, in real-time or large-scale applications, our approach does not rely on querying LLMs during deployment. The inference phase only involves the local model, ensuring that deployment incurs no significant additional computational overhead. Moreover, our method demonstrates superior performance in interpretability compared to the best baseline, achieving enhanced comprehensibility (semantics, distinctiveness, consistency) of 25\%. We believe the limited additional computational cost during training is a worthwhile trade-off for these significant benefits, especially as it does not impact real-world deployment efficiency.

\section{Human Evaluation of using LLMs as Human Proxies}
\label{app:human_verification}
We conducted several human studies to assess the quality of using LLMs for concept comprehensibility evaluation. 
For each task, we presented the concepts extracted using our method to human evaluators and asked them to rate their level of agreement with the conceptual summaries and segment highlightings generated by LLMs. 
For the ratings of agreement with conceptual summaries, we presented all the concepts obtained using our method with three top-activated examples.
For the ratings of agreement with concept-related segment highlightings, we focused on the concepts that LLM identified as having semantic meanings. For each concept, we randomly selected three cases with corresponding LLM-generated highlightings. Each concept was evaluated three times by different workers using a 5-point scale~(1 indicating complete disagreement and 5 indicating complete agreement). We conducted our evaluation on Prolific\footnote{https://app.prolific.com}, a platform for facilitating high-quality human surveys. Each worker is paid \$10.5 per hour and must pass a screening test to take the survey. A total of 42 participants took part in the LLM-generated conceptual summaries evaluation, while 21 participants took part in the LLM-generated segment highlightings evaluation. The demographic information of the evaluators, as provided by Prolific, is shown in Tables~\ref{tab:demographic_summaries} and~\ref{tab:demographic_highlightings}.

The rating results are shown in Table~\ref{tab:argeement_ratings}. Our human study, conducted with diverse participants, confirms that LLMs closely align with human judgment in concept evaluation without introducing much bias. This validates our method of employing LLMs as human proxies for concept evaluation and leveraging their feedback to refine the model.

\begin{table}[H]
\centering
\small
\caption{Ratings of agreement with the LLM-generated conceptual summaries and segment highlightings}
\begin{tabular}{@{}l|ccccccc@{}}
\toprule
 & CEBaB & Beer & Hotel & IMDB & AGnews & Twitter & SciCite \\ \midrule

Summaries & 4.23 & 4.21 & 4.45 & 4.56 & 4.63 & 4.39 & 4.14 \\
Highlightings & 4.56 & 4.24 & 4.40 & 4.62 & 4.42 & 4.33 & 4.18 \\ 

\bottomrule
\end{tabular}
\label{tab:argeement_ratings}
\end{table}

\section{Human Evaluation of Concept Comprehensibility}
\label{app:human_evaluation}

\subsection{Participants Information}
\label{app:participants_info}
Our human evaluation experiments are conducted on Prolific. Each worker is paid \$10.5 per hour and must pass a screening test to take the survey. To ensure more reliable results, each question is evaluated three times by different workers. For the intruder detection and subjective rating experiments, we recruited 42 participants, while 24 participants were recruited for the forward simulatability experiments. The demographic information of the annotators is shown in Tables~\ref{tab:demographic_summaries} and \ref{tab:demographic_local}. 

\subsection{Intruder Detection Experiments Details}
\label{app:intruder_detection}
For the intruder detection experiments, we tested all extracted concepts of each method across seven tasks. Participants were asked to identify the example that did not belong to the same concept from a set of four highlighted examples. Their success rate in identifying the intruder was then calculated. The interface of the task is shown in Fig.~\ref{figs:screencut-1}.

\subsection{Subjective Rating Experiments Details}
\label{app:sub_rating}
For the subjective rating experiments, we present three top-activated examples for each concept along with its summary and ask participants to rate these examples based on the following criteria using a 1-5 scale: 

\noindent\textbf{Consistency}: Do you think this concept formed by all these highlighted text parts has consistent semantic meanings? 

\noindent\textbf{Clarity}: Do you think the semantic meaning of this concept is clear and easy to identify? 

\noindent\textbf{Task Relevance}: Do you think this concept is related to the task? 

\noindent\textbf{Comprehensibility}: Do you think this concept is comprehensible and easy for humans to understand? 

\subsection{Forward Simulatability Experiments Details}
\label{app:forward_simu}
\begin{figure}[h]
    \centering
    \subfigure[Beer]{
    \includegraphics[width=0.34\textwidth]{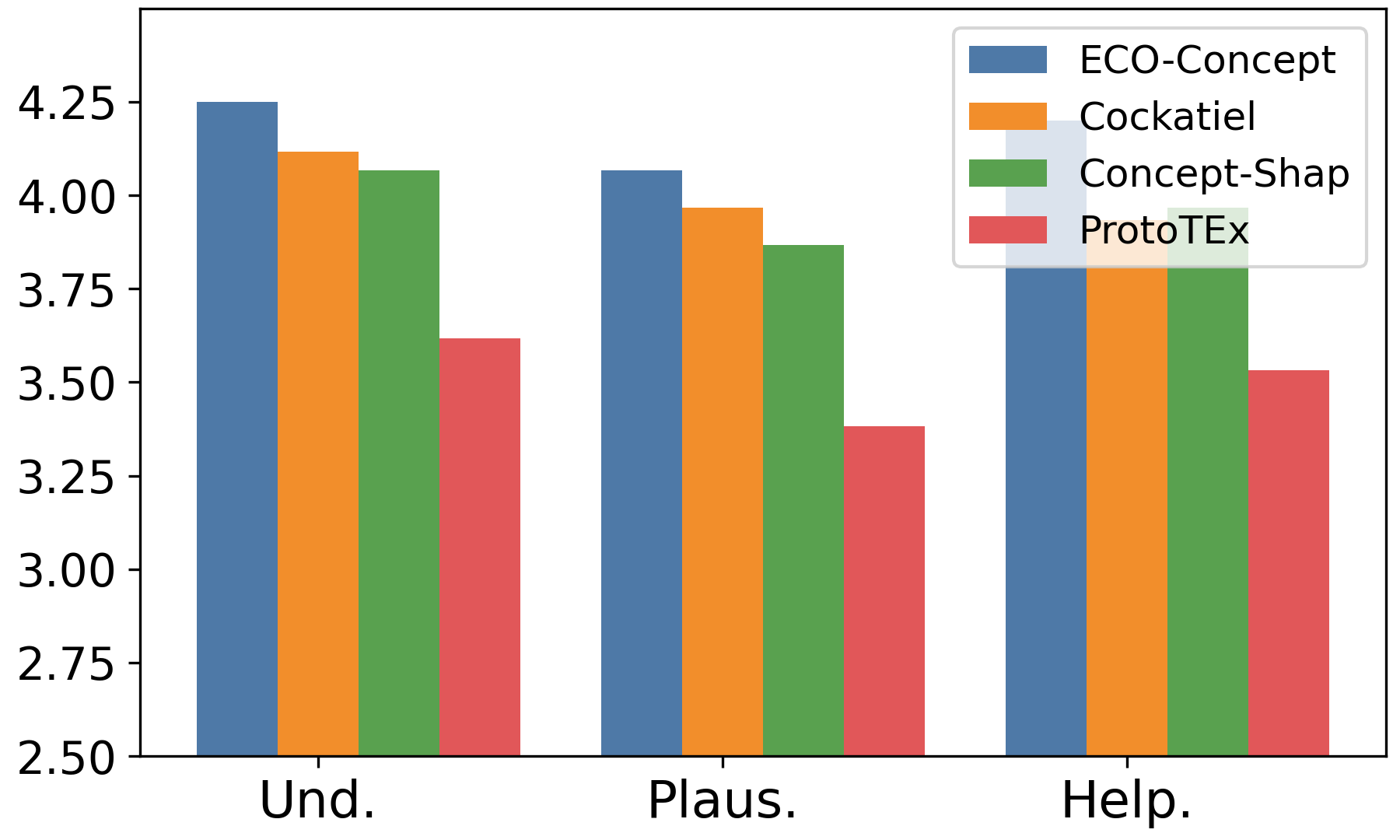}
    }
    \subfigure[AGnews]{
    \includegraphics[width=0.34\textwidth]{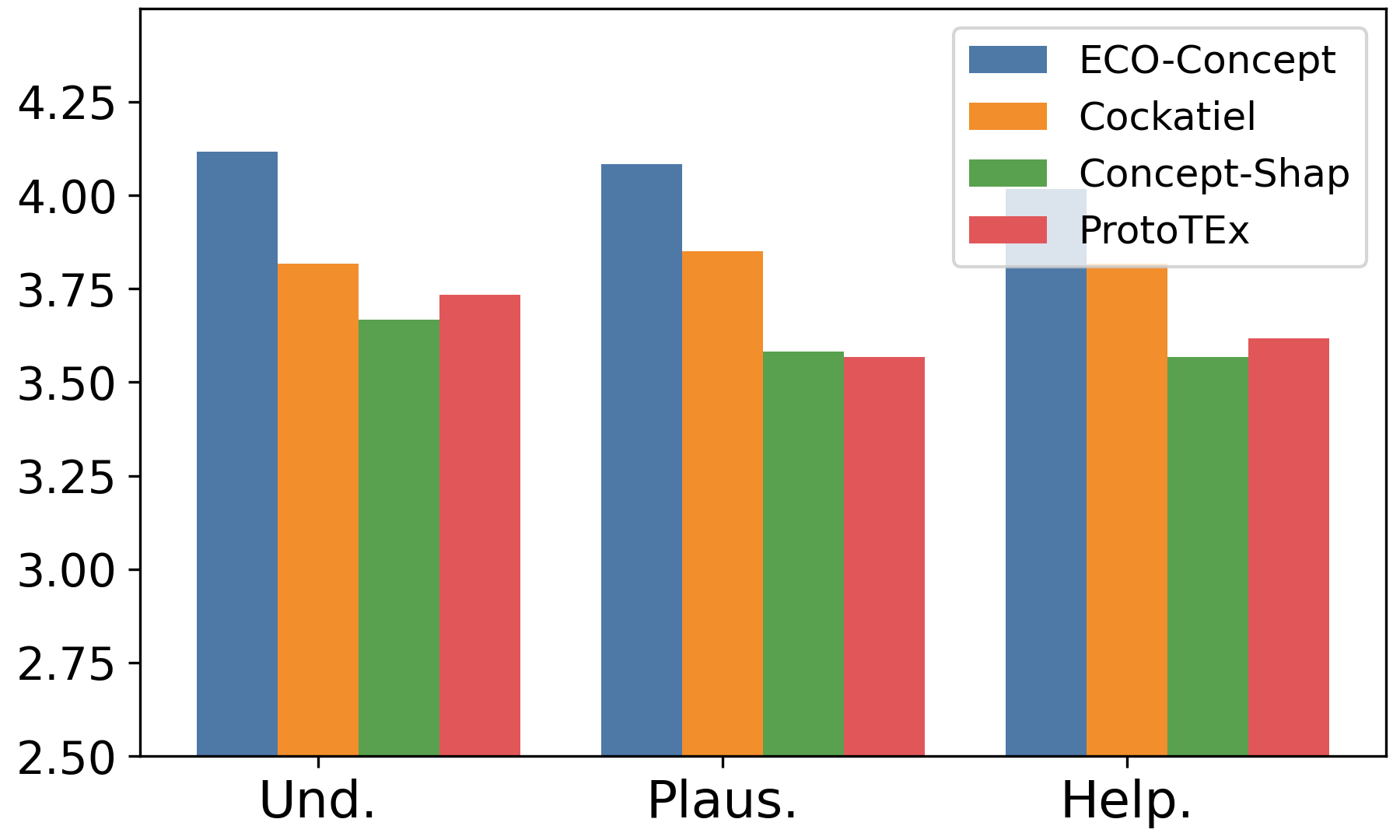}
    }
    \vspace{-0.3cm}
    \caption{Human ratings of provided explanations (Und. indicating understandability, Plaus. indicating plausibility, and Help. indicating helpfulness)}
    \label{figs:human_ratings_local}
\end{figure}

\begin{figure*}[h!]
    \centering
    \subfigure[The survey interface of intruder detection]{
        \includegraphics[width=0.45\textwidth]{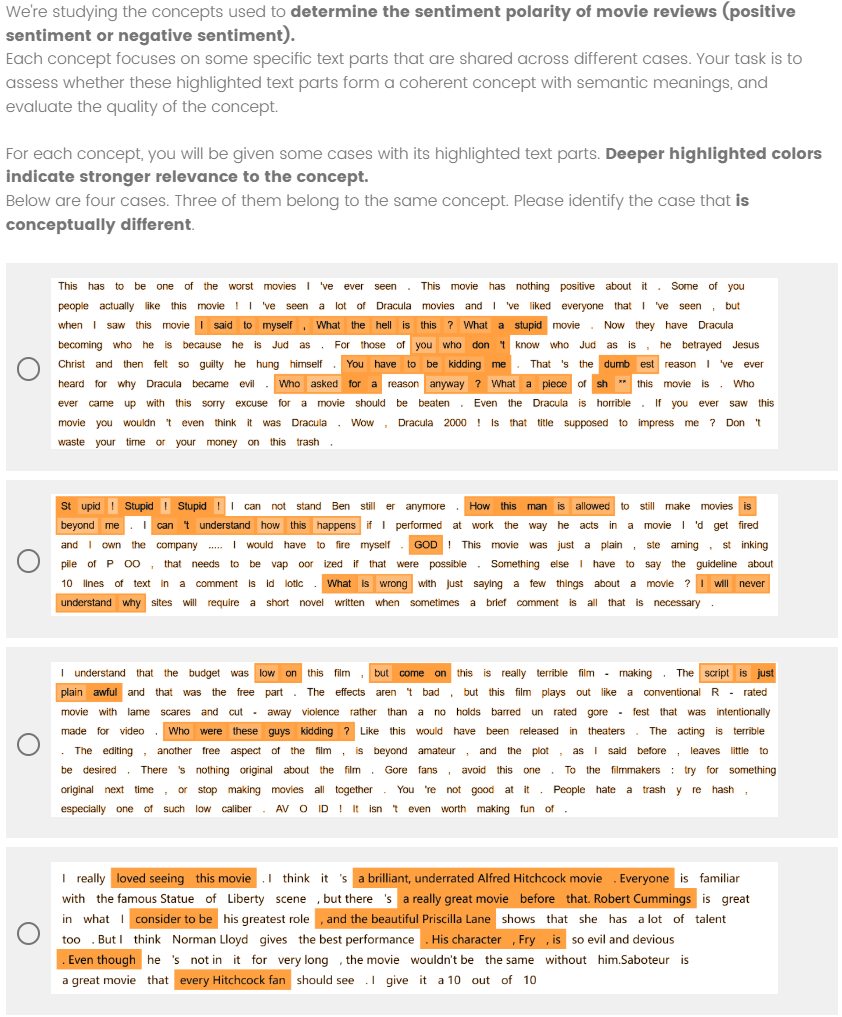}
        \label{figs:screencut-1}
    }
    \subfigure[The tutorial of forward simulatability part~(1)]{
        \includegraphics[width=0.45\textwidth]{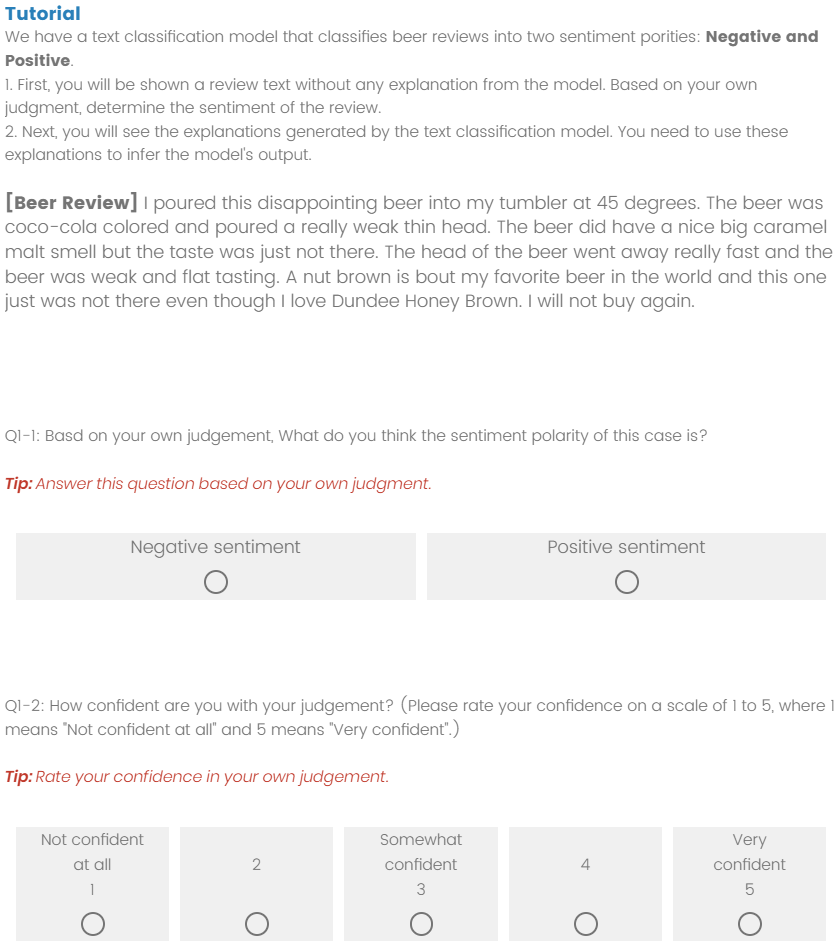}
        \label{figs:screencut-2}
    }
        \subfigure[The tutorial of forward simulatability part~(2)]{
        \includegraphics[width=0.45\textwidth]{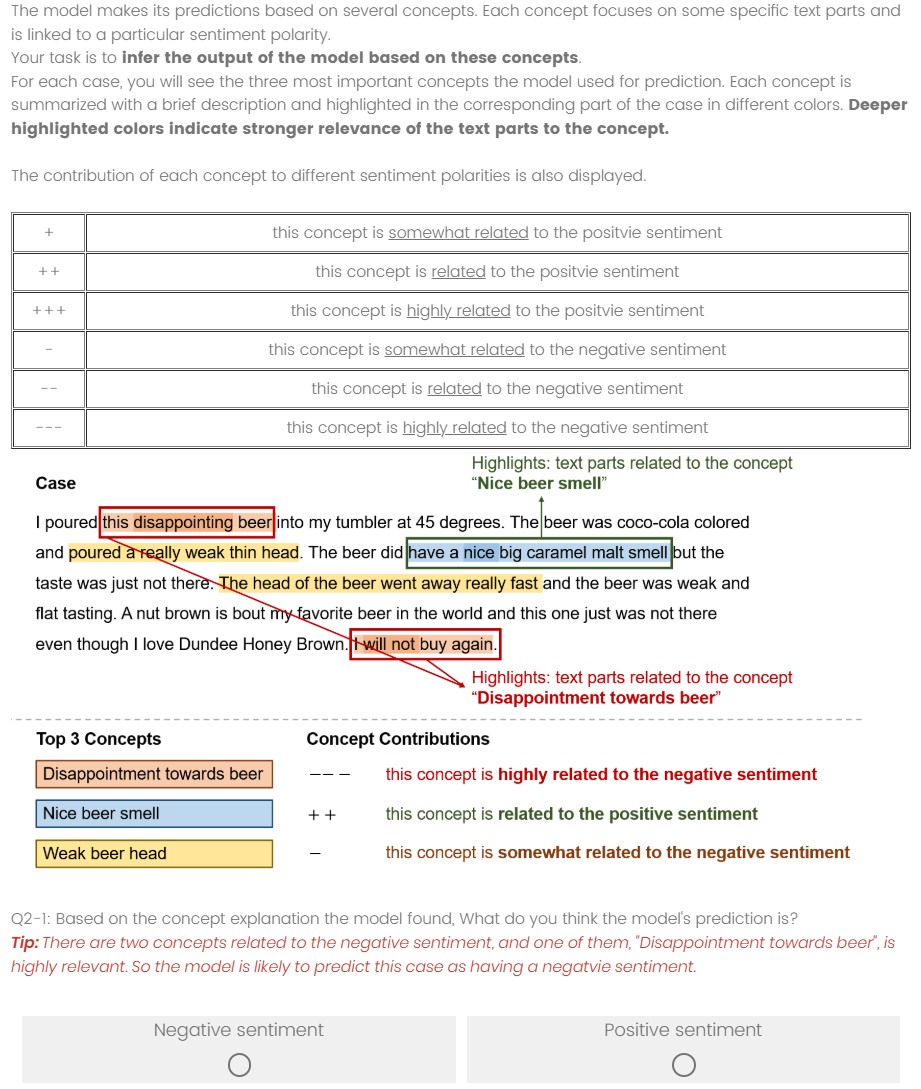}
        \label{figs:screencut-3}
    }
        \subfigure[The tutorial of forward simulatability part~(3)]{
        \includegraphics[width=0.45\textwidth]{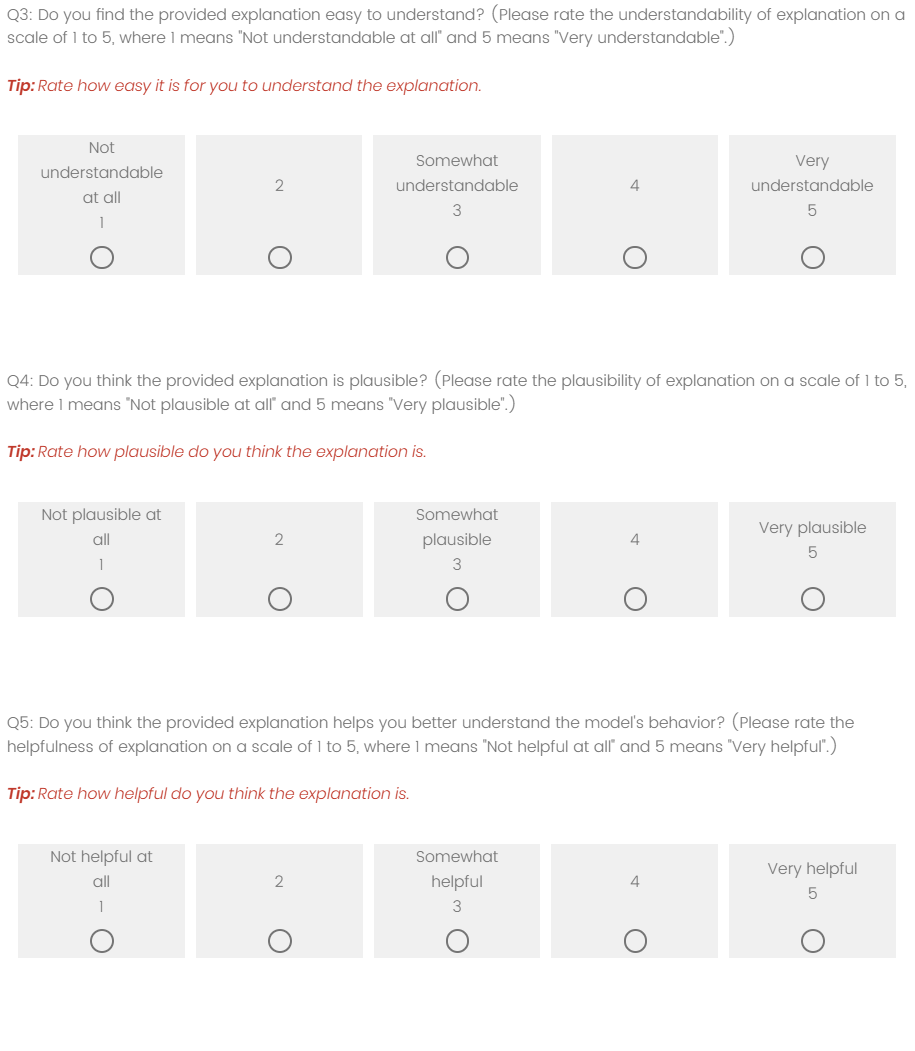}
        \label{figs:screencut-4}
    }
    \caption{Screenshots of the survey interface}
\end{figure*}

For the forward simulatability experiments, we randomly selected 20 samples from each of the Beer and AGnews datasets, ensuring that the accuracy of all methods within these samples was between 80\% and 90\%. First, participants classified the samples without any model explanations. Then, one method’s explanation was provided to the participants, who used this to predict the model’s output and recorded their confidence. To facilitate understanding by non-experts, we normalized the contribution of each concept to the class comparison, as illustrated in Fig.~\ref{figs:screencut-3}. Additionally, participants were asked to rate the understandability, plausibility, and helpfulness of the provided explanation, as Fig.~\ref{figs:screencut-4} shows. The results of the human ratings are presented in Fig.~\ref{figs:human_ratings_local}.

\section{Further Analysis on Ablation Study}
\label{app:ablation}

\subsection{Further Analysis on Impact of Concept Comprehensibility Enhancement}
\label{app:impact_of_e}
\begin{figure}[H]
    \centering
    \subfigure[Explanations provided by the Base model]{
    \includegraphics[width=0.34\textwidth]{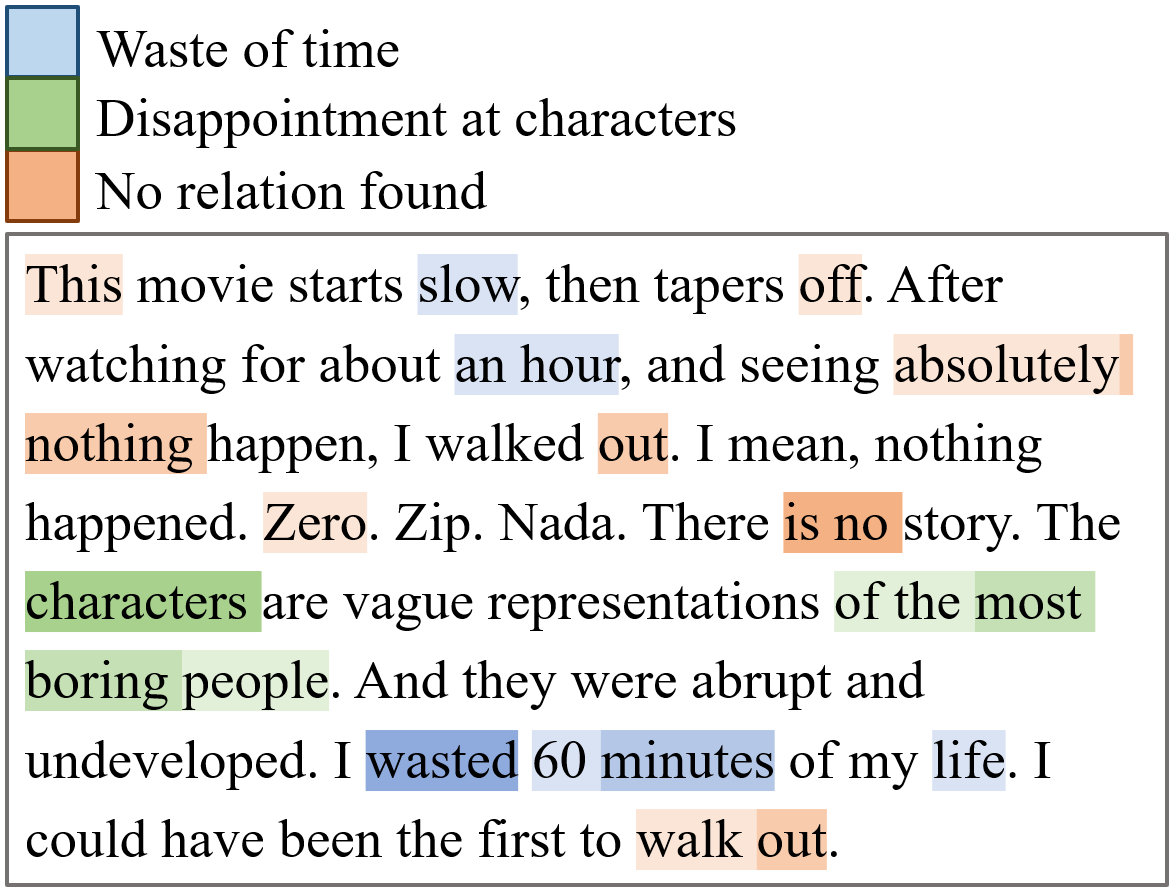}
    }
    \subfigure[Explanations provided by ECO-Concept]{
    \includegraphics[width=0.34\textwidth]{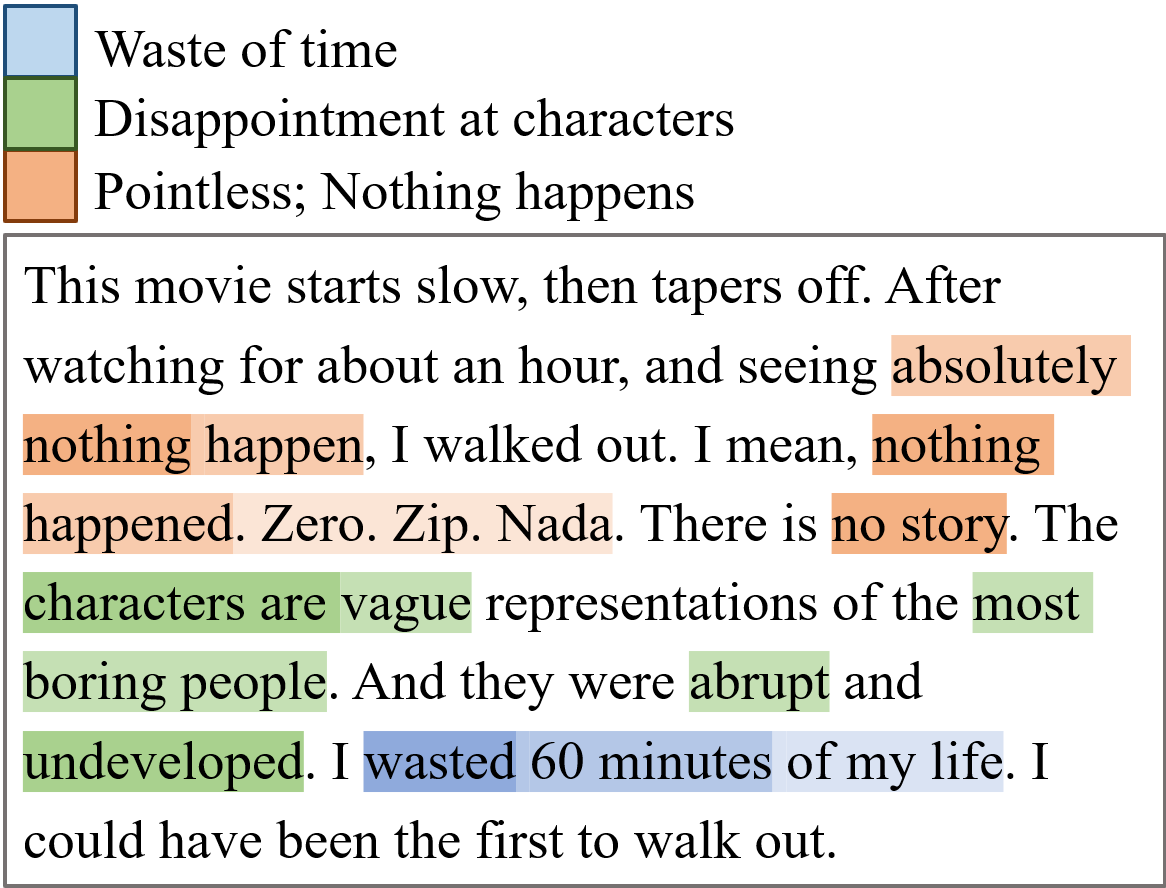}
    }
    \vspace{-0.3cm}
    \caption{Comparison of concept explanations before and after concept enhancement in a beer review case}
    \label{figs:enhance_case}
\end{figure}

As shown in Fig.~\ref{figs:impact_of_e}, for concept comprehensibility metrics, fine-tuning with the comprehensibility loss performs better than fine-tuning without this loss. Moreover, in certain tasks, solely relying on concept regularizers for retraining can even result in a decline in the semantics, distinctiveness, and consistency of the concepts. The possible explanation is that concept regularizers guide concept learning within the tensor space. During the early stages of training, they enhance the consistency of representations for the same concept while improving the distinctiveness between different concepts. As training progresses, most concepts gradually acquire certain semantic meanings. However, without concept annotation guidance, semantic ambiguity becomes unavoidable. This semantic ambiguity is often difficult to resolve in high-dimensional tensor spaces. For semantically ambiguous concepts, some samples may exhibit inconsistencies in semantics, while their features remain relatively close within the tensor space. In such cases, continuing to rely solely on concept regularizers not only fails to resolve the ambiguity but may exacerbate it, thereby compromising the performance of concept metrics. To address this, we propose an enhancement method inspired by the human evaluation process. By incorporating an evaluation mechanism that aligns more closely with human cognition, we can reduce the semantic ambiguity of concepts and enhance their comprehensibility. The above analysis demonstrates the significance and effectiveness of our proposed comprehensibility enhancement stage.
A case in Fig.~\ref{figs:enhance_case} presents that after concept enhancement, the model not only achieves higher semantic consistency but also clarifies the meaning of concepts that were previously semantically ambiguous.

\section{Parameter Sensitivity Analysis}
\label{app:sensitivity}
\subsection{Impact of the Number of Concepts}
\begin{figure*}[h]
    \centering
    \subfigure[CEBaB]{
    \includegraphics[width=0.23\textwidth]{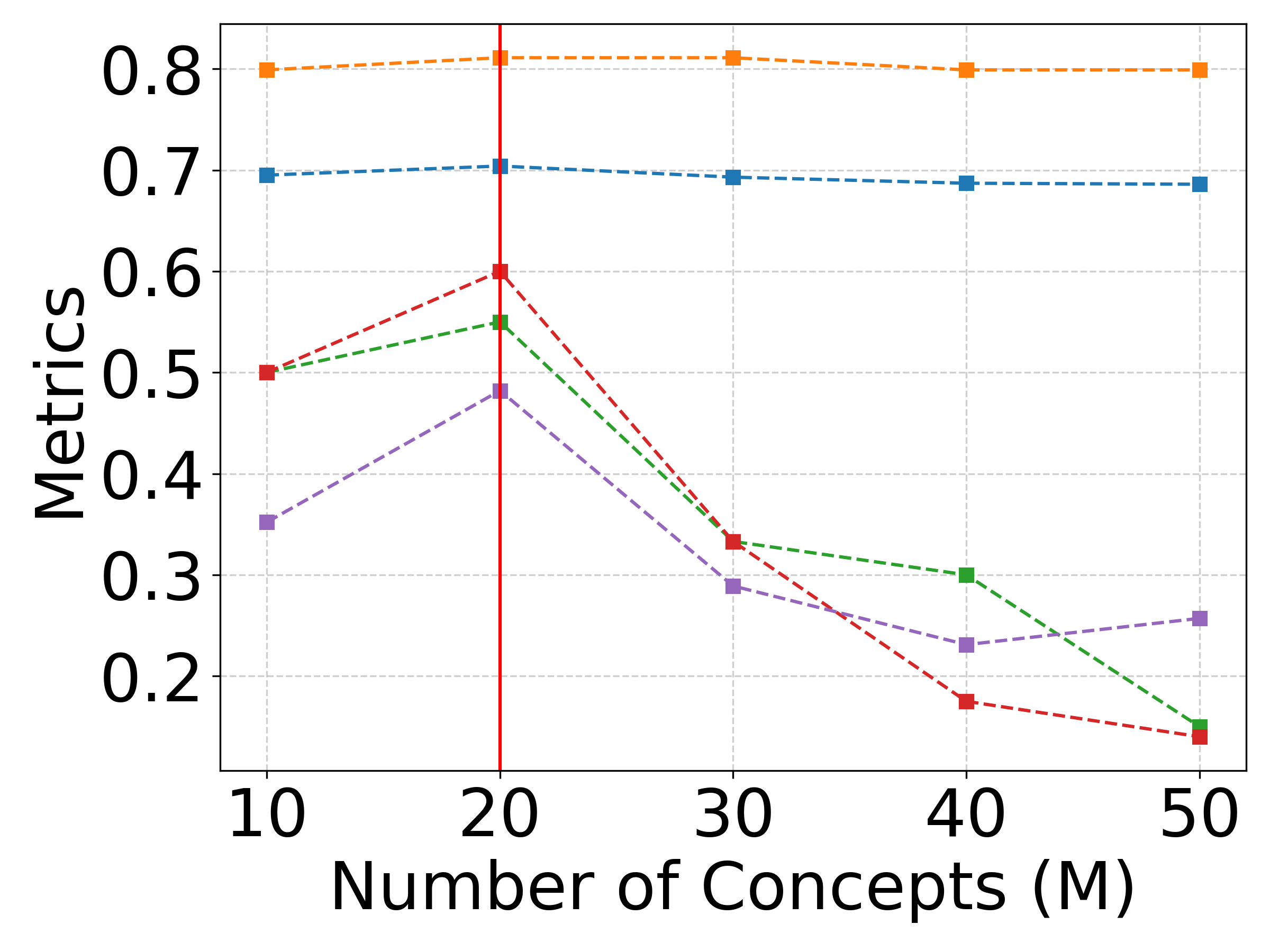}
    }
    \subfigure[Beer]{
    \includegraphics[width=0.23\textwidth]{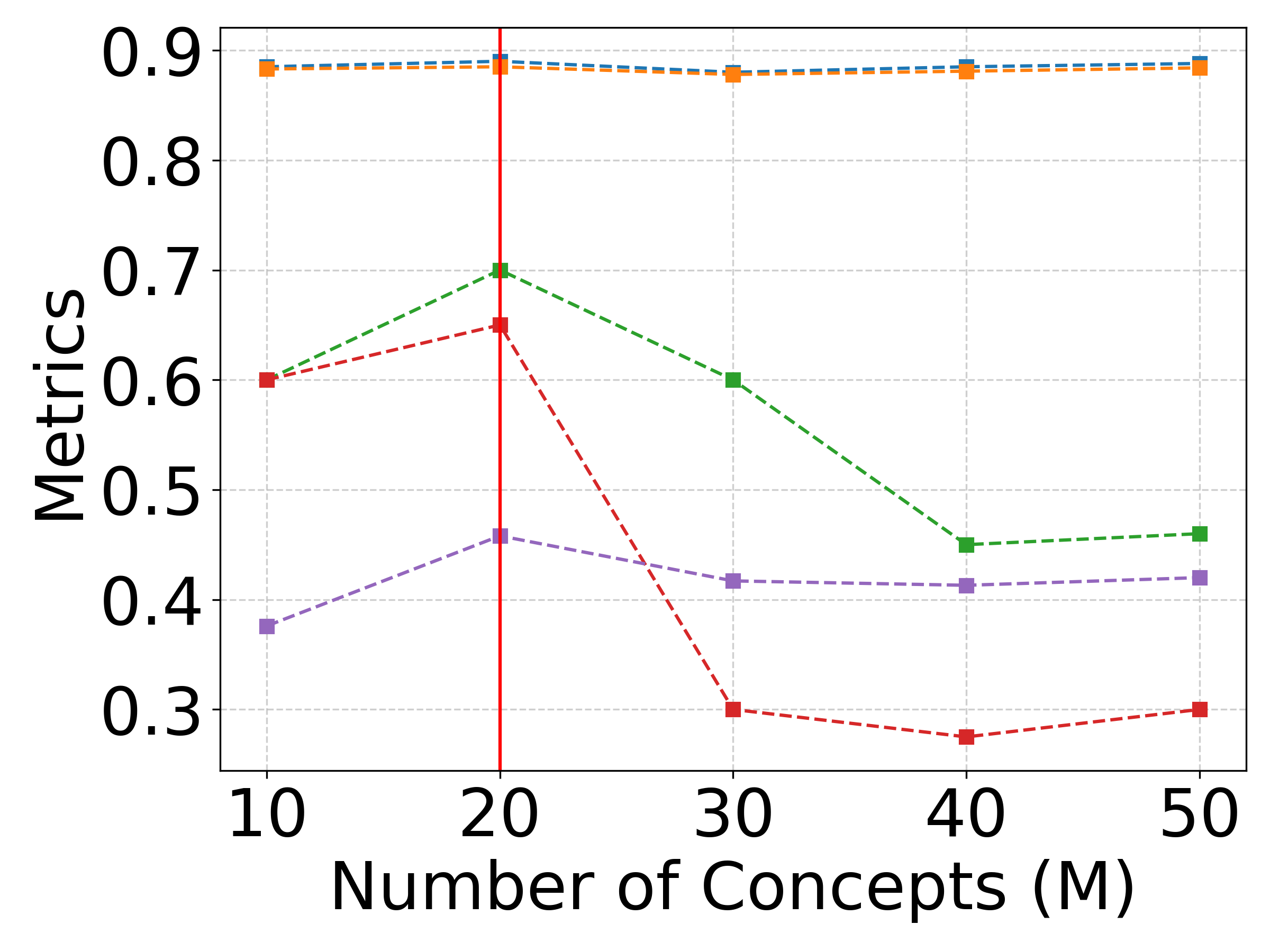}
    }
    \subfigure[Hotel]{
    \includegraphics[width=0.23\textwidth]{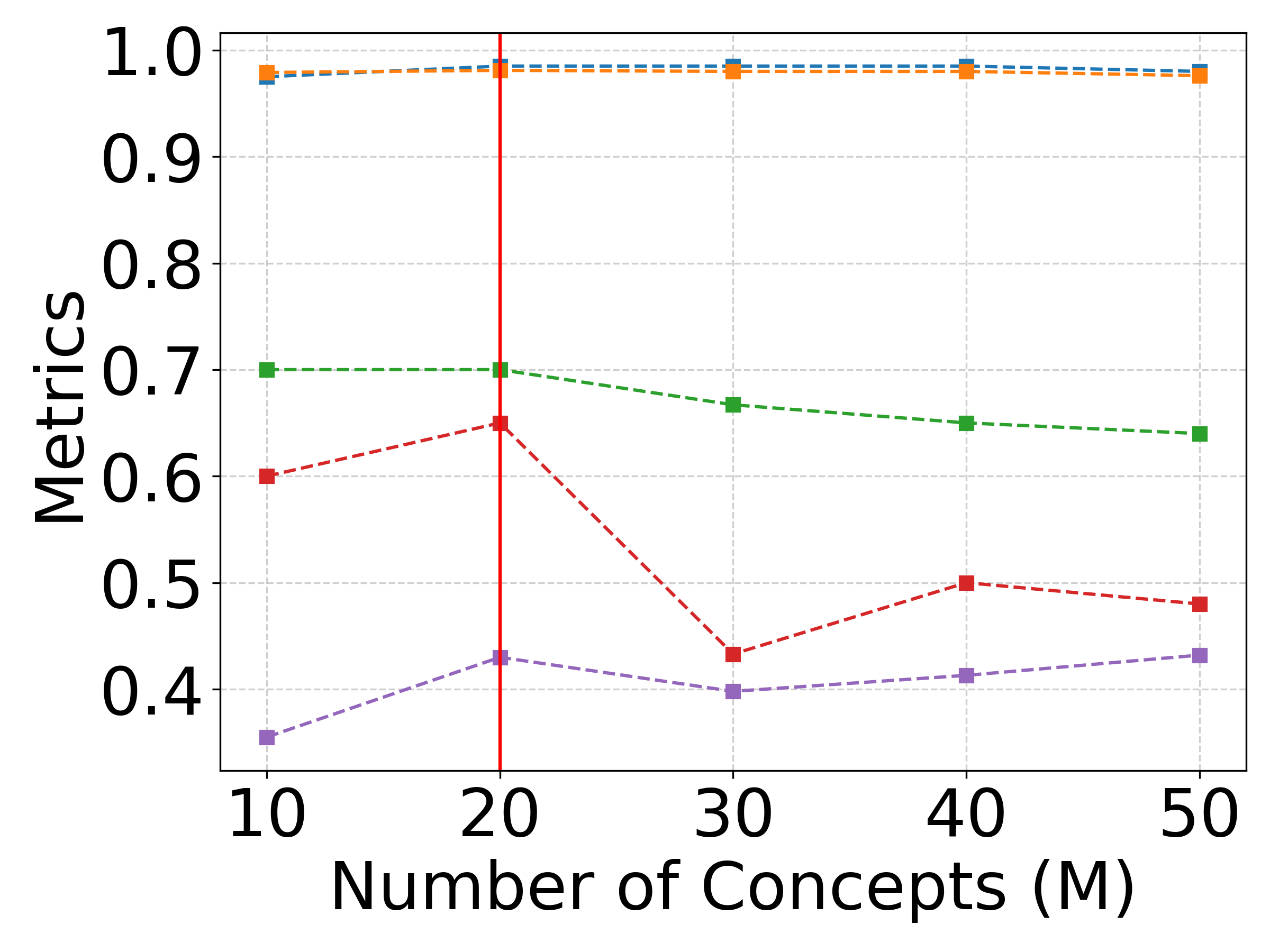}
    }
    \subfigure[IMDB]{
    \includegraphics[width=0.23\textwidth]{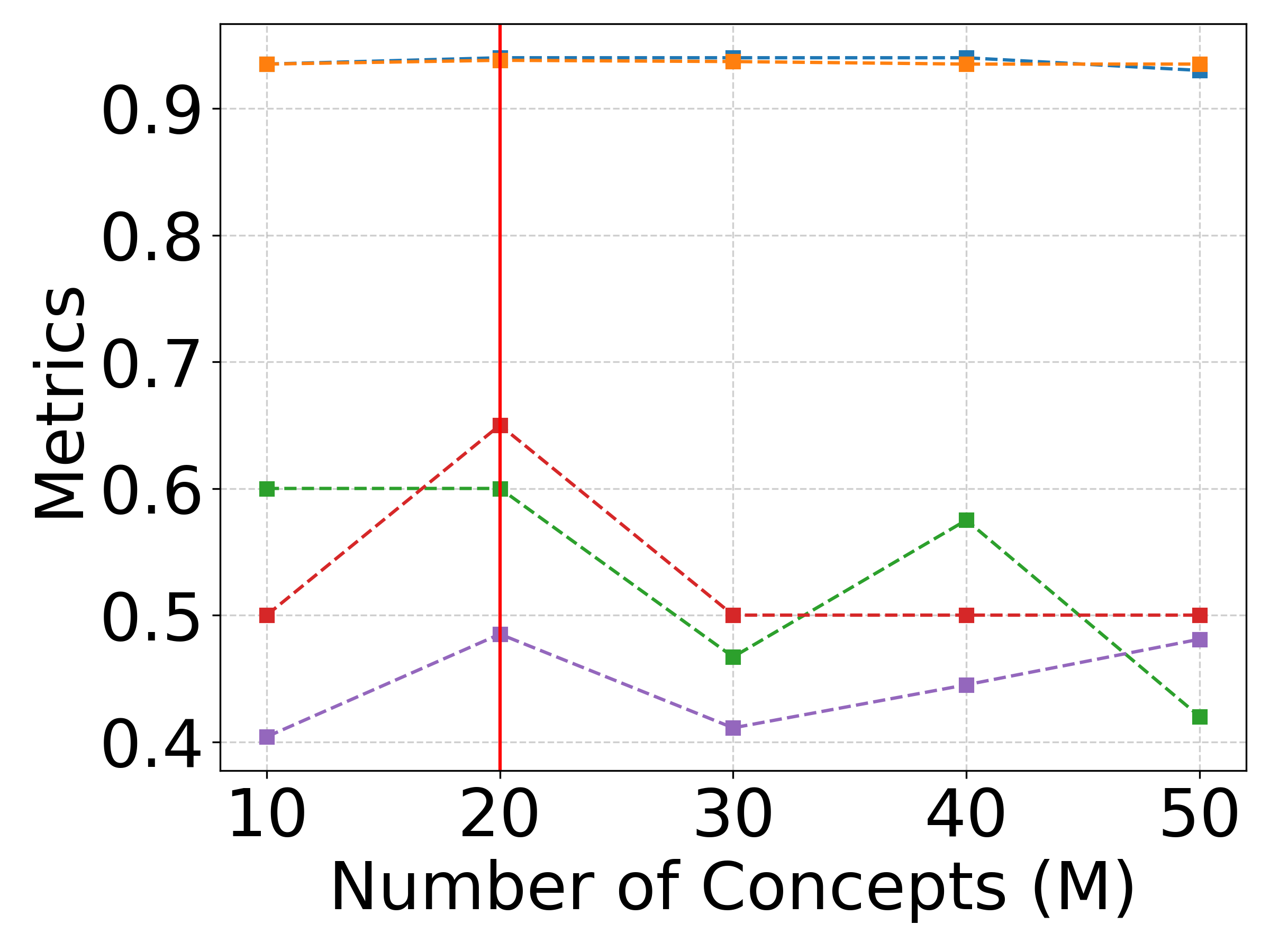}
    }
    \subfigure[AGnews]{
    \includegraphics[width=0.23\textwidth]{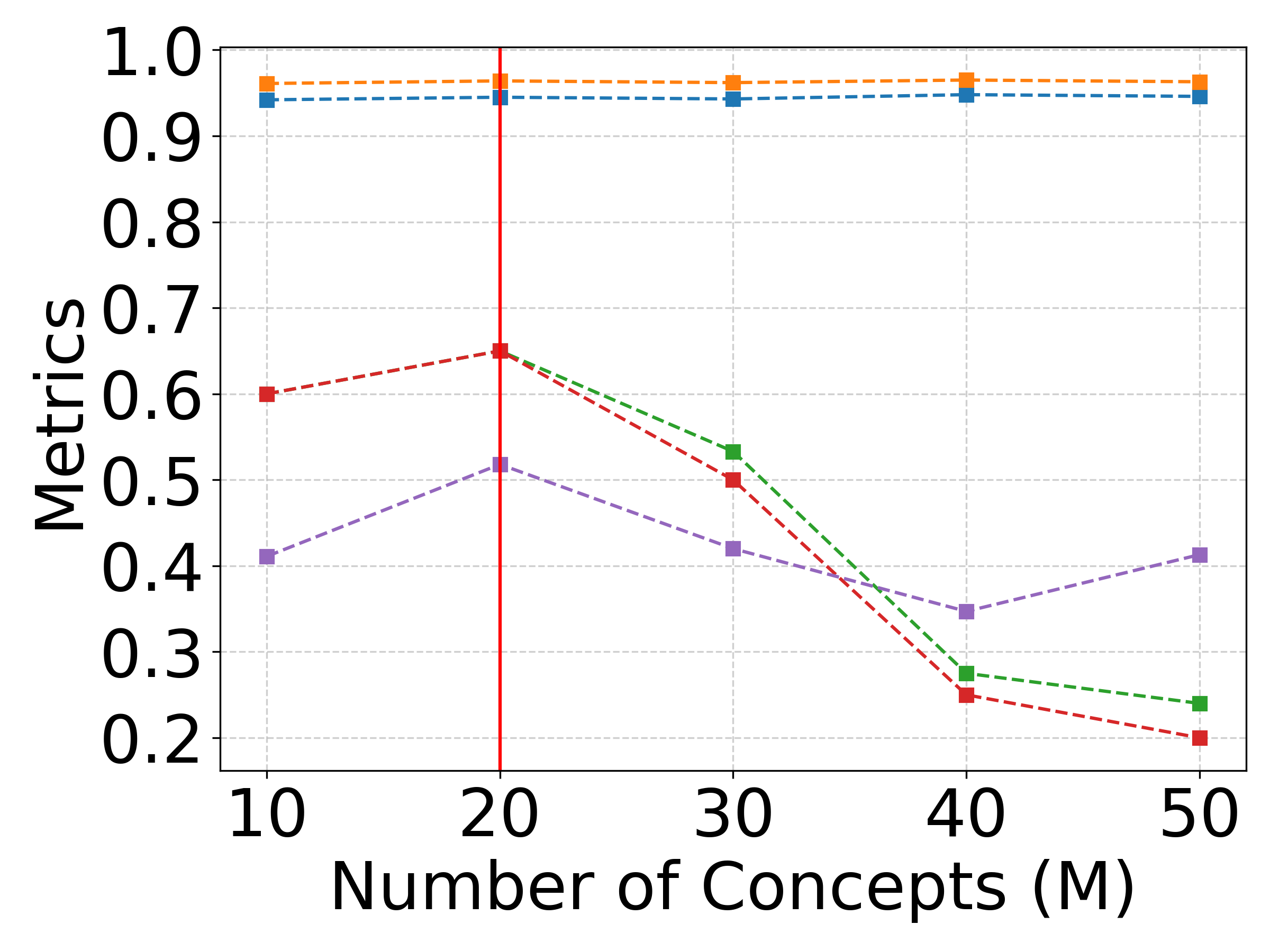}
    }
    \subfigure[Twitter]{
    \includegraphics[width=0.23\textwidth]{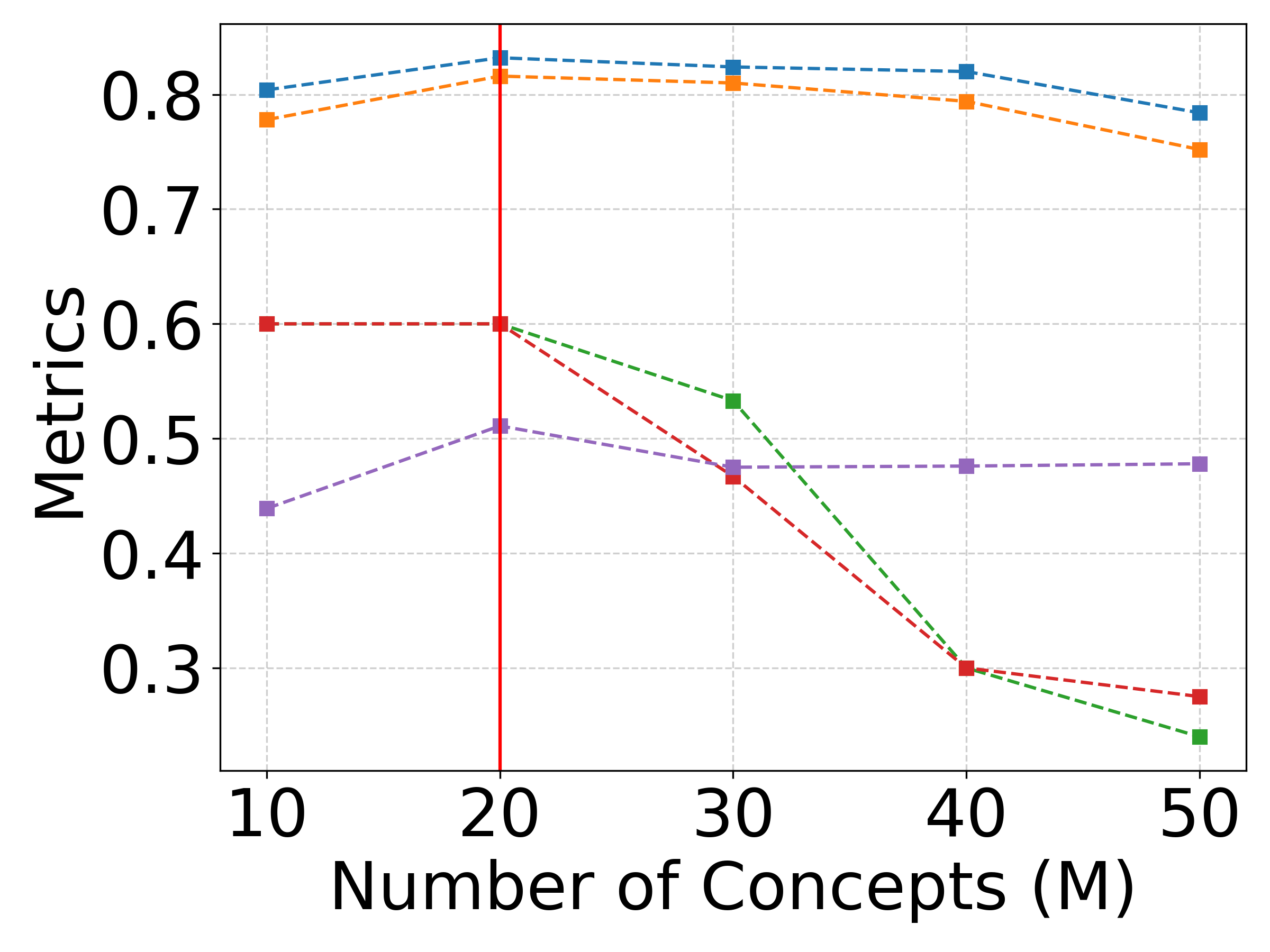}
    }
    \subfigure[SciCite]{
    \includegraphics[width=0.23\textwidth]{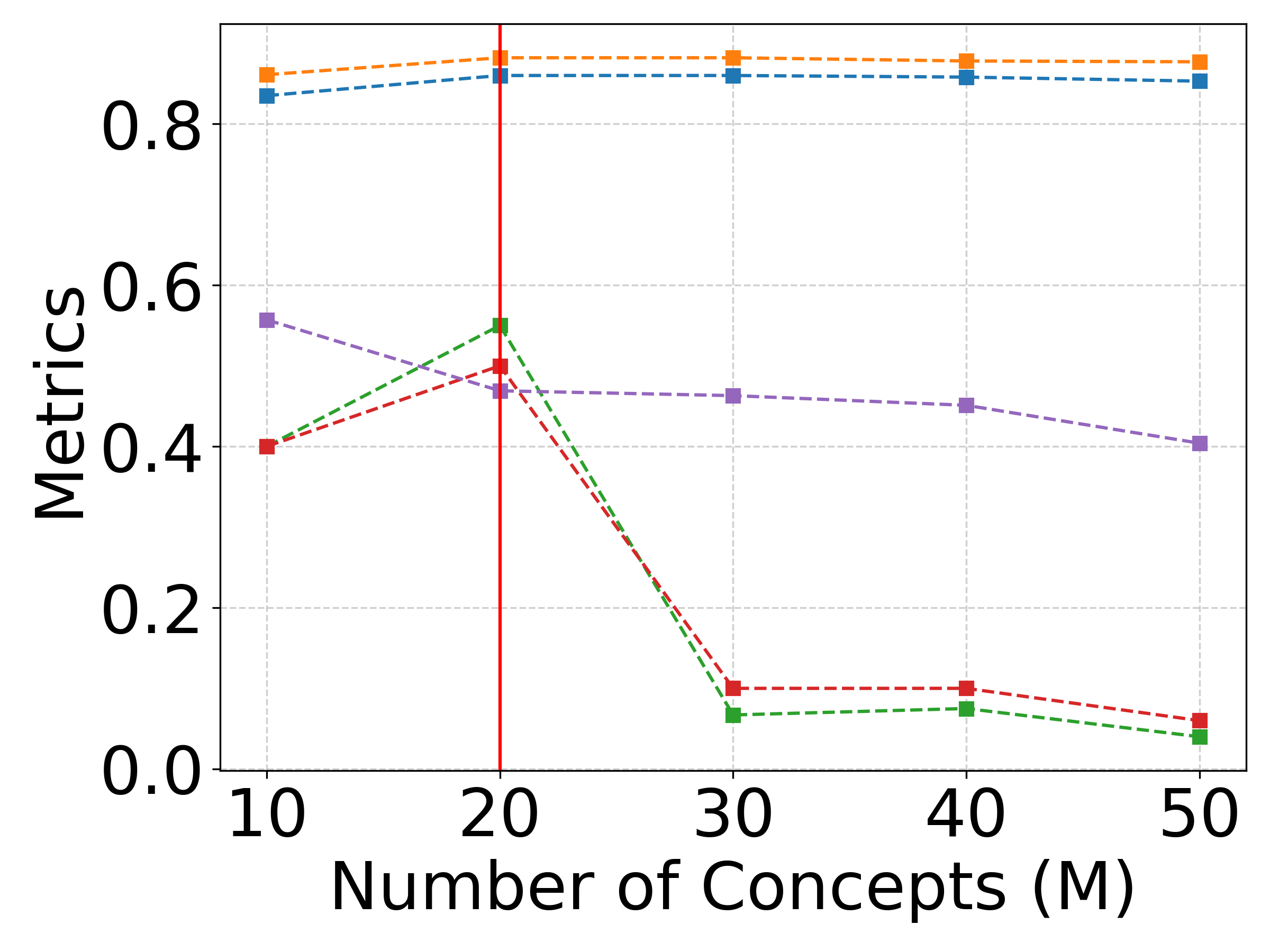}
    } 
    \subfigure[Legend]{
    \includegraphics[width=0.23\textwidth]
    {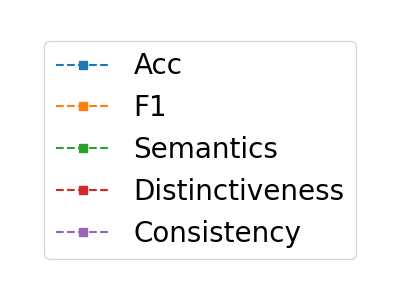}
    }
    \caption{Classification performance and concept metrics of our method with different concept numbers.}
    \label{figs:impact_of_m}
\end{figure*}

To evaluate the impact of the number of concepts on task performance and concept-related metrics, we conducted experiments with 10, 20, 30, 40, and 50 concepts, respectively. The results, as shown in Fig.~\ref{figs:impact_of_m}, indicate that when the number of concepts is 20, most metrics achieve their optimal values. All variants achieve similar results across different tasks, and the number of concepts has minimal effect. For concept interpretability, when the number of concepts exceeds 20, both the concept semantics and distinctiveness decline. This suggests that an excessive number of concepts may bring more semantically irrelevant or redundant concepts. Furthermore, regardless of the number of concepts, it is impossible to ensure that all extracted concepts have semantic meanings due to the unsupervised nature of our method. Nevertheless, when the number of concepts is set to 20, over half of the extracted concepts can be confidently considered semantically meaningful.

\subsection{Grid search Details}

The trade-off parameters $\lambda_{con}$, $\lambda_{dist}$, and $\lambda_{com}$ were selected through grid search, aiming to optimize the model's classification performance on the validation set. The detailed information about the grid search range is presented in Table~\ref{tab:grid_search}.
\begin{table}[H]
\centering
\small
\caption{Details of the grid search range for optimal trade-off parameters}
\resizebox{0.6\textwidth}{!}{
\begin{tabular}{@{}c|cc@{}}
\toprule
Parameter & Search Range & Optimal Value \\ \midrule

$\lambda_{con}$ & 0.01 / 0.05 / 0.1 / 0.3 / 0.5  & 0.1  \\
$\lambda_{dist}$ & -0.001 / -0.005 / -0.01 / -0.05 / -0.1 & -0.01  \\
$\lambda_{com}$ & 0.1 / 0.5 / 1.0 / 1.5 / 2.0 & 1.0  \\

\bottomrule
\end{tabular}}
\label{tab:grid_search}
\end{table}

Our grid-search results indicated that other parameter combinations had lower performance compared to the optimal combination. Table~\ref{tab:grid_search_results} shows the classification performances of the top 4 performing parameter combinations across all tasks.
\begin{table*}[ht]
\centering
\caption{Classification performances of the top 4 performing parameter combinations ($\lambda_{con}$, $\lambda_{dist}$, and $\lambda_{com}$) across all tasks. The best results are bolded.}
\resizebox{\textwidth}{!}{
\begin{tabular}{@{}c| cc cc cc cc cc cc cc@{}}
\toprule
 \multirow{2}{*}{\textbf{Parameter Combination}} & \multicolumn{2}{c}{\textbf{CEBaB}} & \multicolumn{2}{c}{\textbf{Beer}} & \multicolumn{2}{c}{\textbf{Hotel}} & \multicolumn{2}{c}{\textbf{IMDB}} & \multicolumn{2}{c}{\textbf{AGnews}} & \multicolumn{2}{c}{\textbf{Twitter}} & \multicolumn{2}{c}{\textbf{SciCite}} \\
\cmidrule(lr){2-3} \cmidrule(lr){4-5} \cmidrule(lr){6-7} \cmidrule(lr){8-9} \cmidrule(lr){10-11} \cmidrule(lr){12-13} \cmidrule(lr){14-15}
& Acc & F1 & Acc & F1 & Acc & F1 & Acc & F1 & Acc & F1 & Acc & F1 & Acc & F1 \\
\midrule

(0.1, -0.01, 1) & \textbf{.697}
& \textbf{.808}
& \textbf{.885}
& \textbf{.885}
& \textbf{.981}
& \textbf{.981}
& \textbf{.937}
& \textbf{.937}
& \textbf{.941}
& \textbf{.961}
& \textbf{.828}
& \textbf{.813}
& \textbf{.860}
& \textbf{.881} \\

(0.05, -0.01, 1) & .677
& .793
& .882
& .882
& .977
& .977
& .934
& .934
& .927
& .939
& .824
& .801
& .856
& .879 \\

(0.1, -0.01, 0.5) & .683
& .799
& .883
& .883
& .979
& .979
& .931
& .931
& .936
& .951
& .826
& .804
& .856
& .879 \\

(0.05, -0.01, 0.5) 
& .686
& .799
& .883
& .883
& .977
& .977
& .929
& .929
& .929
& .948
& .823
& .799
& .858
& .880\\

\bottomrule
\end{tabular}}
\label{tab:grid_search_results}
\vspace{-0.2cm}
\end{table*}

\section{Model Robustness Evaluation}
\label{app:robustness}
To test whether our method can yield reliable explanations for noisy samples, we performed experiments using seven datasets. We randomly chose 100 test samples and applied noisy perturbations to 5\% of the words in each sample. These words were chosen based on their frequency in the training set (excluding stop words), with higher-frequency words being more likely to be perturbed. For each sample, we randomly applied one of these adversarial strategies: synonym replacement or spelling error. For synonym replacement, we selected a synonym from WordNet~\citep{miller1995wordnet}, prioritizing the one with the lowest frequency in the training set. For spelling errors, we randomly altered a single character in the word.

To evaluate the robustness of our model's explanations, we compared the top concepts attributed to the same sample before and after the noisy perturbations. We used Kendall’s rank correlation~\citep{kendall1938new} as the evaluation metric, which accounts for the concept importance rankings. The rank correlation scores, calculated based on the ranks of the top K attributed concepts before and after noisy perturbations, are shown in Table~\ref{tab:robustness}. After noisy perturbations, our model is still able to attribute the test samples to concept explanations that closely match those before the attack. This demonstrates the robustness of the explanations provided by our method under noisy conditions, highlighting its potential for practical applications. 

\begin{table}[H]
\centering
\small
\caption{Rank correlation scores calculated based on the ranks of the top K
attributed concepts before and after noisy perturbations}
\begin{tabular}{@{}l|ccccccc@{}}
\toprule
 & CEBaB & Beer & Hotel & IMDB & AGnews & Twitter & SciCite \\ \midrule

K=3 & .895 & .925 & .934 & .947 & .897 & .916 & .919 \\
K=5 & .817 & .826 & .803 & .821 & .802 & .844 & .870 \\ 

\bottomrule
\end{tabular}
\label{tab:robustness}
\end{table}

\section{Prompts}
The prompts we used are shown in Table~\ref{tab:sum_prompts}.

\section{Comparison of Explanations}
\begin{figure*}[h]
    \centering
    \subfigure[ECO-Concept]{
    \includegraphics[width=0.9\textwidth]{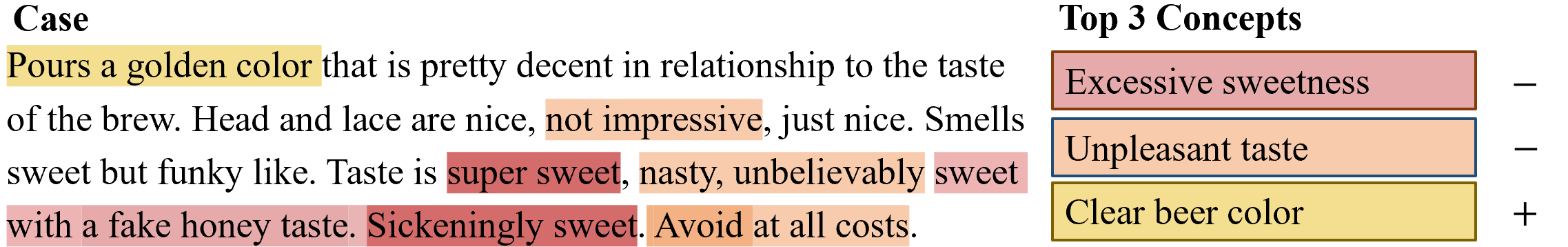}
    }
    \subfigure[Cockatiel]{
    \includegraphics[width=0.9\textwidth]{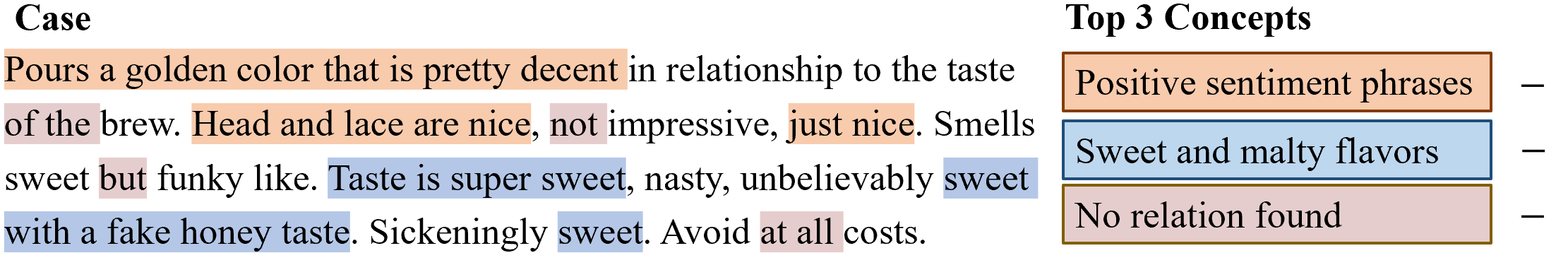}
    }
    \subfigure[Concept-Shap]{
    \includegraphics[width=0.9\textwidth]{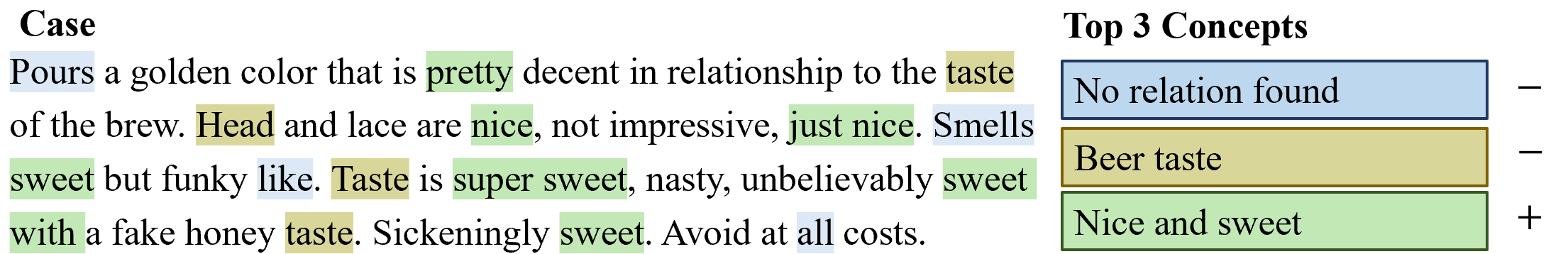}
    }
    \subfigure[ProtoTEx]{
    \includegraphics[width=0.9\textwidth]{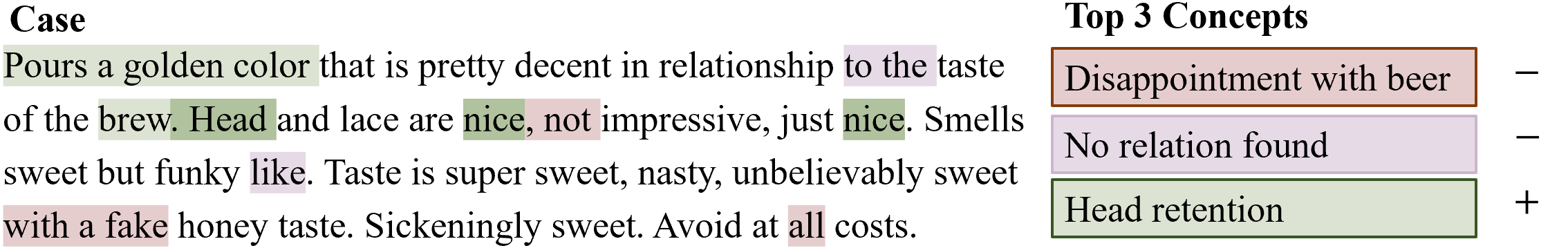}
    }
    \caption{Comparison of concept explanations provided by ECO-Concept and other concept-based methods}
    \label{figs:explanation_cases}
\end{figure*}

The comparison of explanations of different concept-based methods is shown in Fig.~\ref{figs:explanation_cases}.

\begin{table*}[h]
\centering
\small
\caption{Prompt for concept summary and concept-related segment highlightings in the movie review sentiment classification task}
\begin{tabular}{p{\textwidth}} 
\toprule
\rowcolor[gray]{0.9}
\textit{Prompt for Concept Summary} \\ 
\midrule
\textbf{Prompt:} We're studying the concepts used to determine whether the sentiment of movie reviews is positive or negative. Each concept focuses on some specific elements. First, pay attention to all highly activated text parts in the following example sentences. Then think deeply to find the relations between all the highly activated text parts. Finally, determine whether all these highly activated text parts can represent a consistent concept. If not, please output 'No relation found'; if yes, provide a summary in no more than ten words. The activation format is token<tab>activation. Activation values range from 0 to 1. A concept finding what it's looking for is represented by a non-zero activation value. The higher the activation value, the stronger the match.\\ 
\end{tabular}
\begin{tabular}{p{\textwidth}} 
\toprule
\rowcolor[gray]{0.9}
\textit{Prompt for Concept-related segment highlightings} \\ 
\midrule
\textbf{Prompt:} We're studying the concepts used to determine whether the sentiment of movie reviews is positive or negative. Each concept focuses on specific elements in a short document. Look at the summary of what the concept is, and try to determine whether each token in the document is related to the concept. Return 1 for related tokens and 0 for unrelated tokens. You should carefully consider whether each token is related to the concept and not return 1 for all the tokens in the document. \\ 
\bottomrule
\end{tabular}
\label{tab:sum_prompts}
\end{table*}

\begin{table*}[htbp]
    \centering
    \small
    \caption{Demographic information of participants in LLM-generated conceptual summaries evaluation, intruder detection experiments, and subjective rating experiments}
    \begin{tabular}{>{\centering\arraybackslash} p{2.2cm}>{\centering\arraybackslash}p{2.2cm}>{\centering\arraybackslash}p{2.2cm}>{\centering\arraybackslash}p{2.2cm}>{\centering\arraybackslash}p{2.2cm}}
\toprule
        \multicolumn{4}{c}{Highest education level completed} & Avg. Age \\
        \midrule
        High School & Undergraduate & Graduate & Doctorate &  \\
        5 & 21 & 11 & 5 & 36.4 \\
    \end{tabular}
    \begin{tabular}{>{\centering\arraybackslash} p{2.2cm}>{\centering\arraybackslash}p{2.2cm}>{\centering\arraybackslash}p{2.2cm}>{\centering\arraybackslash}p{2.2cm}>{\centering\arraybackslash}p{2.2cm}}
        \toprule
        \multicolumn{4}{c}{Ethnicity} &  Female:Male \\
        \midrule
        White & Black & Asian & Mixed &  \\
        21 & 12 & 5 & 4 & 1:1.62 \\
        \bottomrule
    \end{tabular}
\label{tab:demographic_summaries}
\end{table*}

\begin{table*}[htbp]
    \centering
    \small
    \caption{Demographic information of participants in LLM-generated concept-related highlightings evaluation}
    \begin{tabular}{>{\centering\arraybackslash} p{2.2cm}>{\centering\arraybackslash}p{2.2cm}>{\centering\arraybackslash}p{2.2cm}>{\centering\arraybackslash}p{2.2cm}>{\centering\arraybackslash}p{2.2cm}}
        \toprule
        \multicolumn{4}{c}{Highest education level completed} & Avg. Age \\
        \midrule
        High School & Undergraduate & Graduate & Doctorate &  \\
        3 & 10 & 7 & 1 & 29.1 \\
    \end{tabular}
    \begin{tabular}{>{\centering\arraybackslash} p{2.2cm}>{\centering\arraybackslash}p{2.2cm}>{\centering\arraybackslash}p{2.2cm}>{\centering\arraybackslash}p{2.2cm}>{\centering\arraybackslash}p{2.2cm}}
        \toprule
        \multicolumn{4}{c}{Ethnicity} &  Female:Male \\
        \midrule
        White & Black & Asian & Mixed &  \\
        8 & 6 & 5 & 2 & 1:1.33 \\
        \bottomrule
    \end{tabular}
\label{tab:demographic_highlightings}
\end{table*}

\begin{table*}[htbp]
    \centering
    \small
    \caption{Demographic information of participants in forward simulatability experiments}
    \begin{tabular}{>{\centering\arraybackslash} p{2.2cm}>{\centering\arraybackslash}p{2.2cm}>{\centering\arraybackslash}p{2.2cm}>{\centering\arraybackslash}p{2.2cm}>{\centering\arraybackslash}p{2.2cm}}
        \toprule
        \multicolumn{4}{c}{Highest education level completed} & Avg. Age \\
        \midrule
        High School & Undergraduate & Graduate & Doctorate &  \\
        3 & 11 & 8 & 2 & 40.2 \\
    \end{tabular}
    \begin{tabular}{>{\centering\arraybackslash} p{2.2cm}>{\centering\arraybackslash}p{2.2cm}>{\centering\arraybackslash}p{2.2cm}>{\centering\arraybackslash}p{2.2cm}>{\centering\arraybackslash}p{2.2cm}}
        \toprule
        \multicolumn{4}{c}{Ethnicity} &  Female:Male \\
        \midrule
        White & Black & Asian & Mixed &  \\
        11 & 6 & 4 & 3 & 1:1.18 \\
        \bottomrule
    \end{tabular}
\label{tab:demographic_local}
\end{table*}


\end{document}